\documentclass[runningheads]{llncs}

\usepackage{eccv}

\usepackage{eccvabbrv}

\usepackage{graphicx}
\usepackage{booktabs}
\usepackage[misc]{ifsym}
\usepackage{lipsum}  
\usepackage[accsupp]{axessibility}  

\usepackage{listings}
\usepackage[ruled,linesnumbered]{algorithm2e}
\SetKwComment{Comment}{/* }{ */}
\usepackage{amsmath}  
\usepackage{multicol}
\usepackage{multirow}
\usepackage{makecell}
\usepackage{adjustbox}
\usepackage{bbding}
\usepackage{appendix}

\newcommand{\boldblue}[1]{\textcolor{Blue}{\textbf{#1}}}
\newcommand{\boldred}[1]{\textcolor{Red}{\textbf{#1}}}
\newcolumntype{T}{@{\hspace{0em}}c@{\hspace{0em}}}

\usepackage[breaklinks=true,
            bookmarks=false,
            colorlinks,
            linkcolor=red,       
            anchorcolor=blue,  
            citecolor=green, 
            ]{hyperref}

\usepackage{orcidlink}

\begin{document}

\title{OmniSSR: Zero-shot Omnidirectional Image Super-Resolution using Stable Diffusion Model} 

\titlerunning{OmniSSR}

\author{
Runyi Li$^{ \kern0.5pt \dagger}$
\and
Xuhan Sheng$^{ \kern0.5pt \dagger}$
\and
Weiqi Li
\and
Jian Zhang \kern-1pt \inst{\textrm{\Letter}}
}

\authorrunning{Runyi Li et al.}

\institute{School of Electronic and Computer Engineering, Peking University}

\maketitle

\newcommand\blfootnote[1]{%
\begingroup
\renewcommand\thefootnote{}\footnote{#1}%
\addtocounter{footnote}{-1}%
\endgroup
}

\begin{abstract}
Omnidirectional images (ODIs) are commonly used in real-world visual tasks, and high-resolution ODIs help improve the performance of related visual tasks. Most existing super-resolution methods for ODIs use end-to-end learning strategies, resulting in inferior realness of generated images and a lack of effective out-of-domain generalization capabilities in training methods. Image generation methods represented by diffusion model provide strong priors for visual tasks and have been proven to be effectively applied to image restoration tasks. Leveraging the image priors of the \textbf{S}table Diffusion (SD) model, we achieve \textbf{omni}directional image \textbf{s}uper-\textbf{r}esolution with both fidelity and realness, dubbed as \textbf{OmniSSR}. Firstly, we transform the equirectangular projection (ERP) images into tangent projection (TP) images, whose distribution approximates the planar image domain. Then, we use SD to iteratively sample initial high-resolution results. At each denoising iteration, we further correct and update the initial results using the proposed Octadecaplex Tangent Information Interaction (OTII) and Gradient Decomposition (GD) technique to ensure better consistency. Finally, the TP images are transformed back to obtain the final high-resolution results. Our method is zero-shot, requiring no training or fine-tuning. Experiments of our method on two benchmark datasets demonstrate the effectiveness of our proposed method. 
\end{abstract}

\blfootnote{$\dagger$ means equal contribution.}
\blfootnote{\textrm{\Letter} means corresponding author.}

\section{Introduction}
\label{sec:intro}


Omnidirectional images (ODIs) capture the entire scene in all directions, exceeding the narrow field of view (FOV) offered by planar images. Super-Resolution (SR) techniques enhance the visual quality of ODIs by increasing their resolution, thereby revealing finer details and enabling more accurate scene analysis and interpretation. This becomes particularly crucial in applications like virtual reality and surveillance, where high-resolution ODIs are essential for precise perception and decision-making.

Current research in omnidirectional image super-resolution (ODISR) explores various methodologies to enhance the resolution of ODIs~\cite{ozcinar2019super,deng2022omnidirectional}.
SphereSR~\cite{Yoon_Chung_Wang_Yoon} addresses non-uniformity in different projections by learning upsampling processes and ensuring information consistency using LIIF~\cite{Chen_Liu_Wang_2021}. 
OSRT~\cite{osrt_Yu_Wang_Cao_Li_Shan_Dong_2023} designs a distortion-aware Transformer to modulate equirectangular projection (ERP) distortions continuously and self-adaptively. Without a cumbersome process, OSRT outperforms previous methods remarkably.
However, existing ODISR methods face the following challenges: (1) The majority are end-to-end models that can only produce a deterministic output, always better data fidelity but worse visual perception quality~\cite{duan2018perceptual}. It's promising to develop a generation-based model, but requiring high data demands, yet high-resolution ODIs are high cost to collect~\cite{yagi1999omnidirectional,yamazawa1993omnidirectional}. (2) Most methods directly perform SR on ERP format ODIs, while users usually watch ODIs in a narrow FOV using tangent projection (TP). So another promising direction is to use off-the-shelf planar models on TP images. Recent times have witnessed the introduction and widespread application of diffusion models~\cite{ho2020denoising,song2020score}, especially Stable Diffusion (SD)~\cite{Rombach_Blattmann_Lorenz_Esser_Ommer_2022}, which have provided a robust backbone for visual tasks~\cite{jiang2022reference, yang2023rerender, yu2023cross, guo2024cas}, including SR~\cite{wang2022zero, StableSR_Wang_Yue_Zhou_Chan_Loy_2023, sr3,liu2023residual,Xia_Zhang_Wang_Wang_Wu_Tian_Yang_Gool,yue2024resshift}. However, if TP images are trivially one-by-one processed using diffusion-based SR models, they will exhibit discrepancies in the overlapping region when re-projected onto the ERP image. As a result, the global continuity is compromised.

\begin{figure}
    \centering
    \includegraphics[width=1\linewidth]{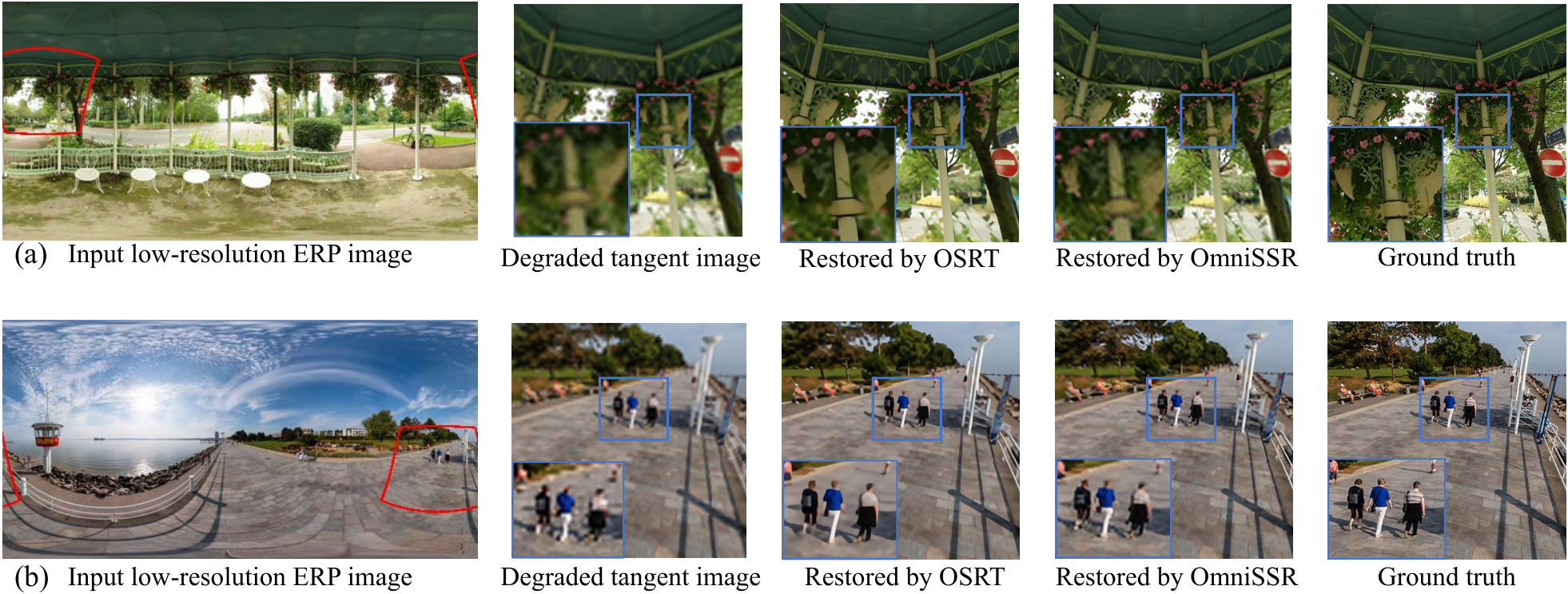}
    \caption{We address omnidirectional image super-resolution in a \textit{zero-shot} manner via OmniSSR. Presented above are select outcomes that sketch the essence of OmniSSR compared with current state-of-the-art approach OSRT~\cite{osrt_Yu_Wang_Cao_Li_Shan_Dong_2023}. Part (a) and (b) illustrate
    that OmniSSR upholds fidelity and visual realness at the same time, providing vivid and realistic details, while OSRT outputs over-smoothed and distorted results. Zoom in for more details.}
    \label{fig:teaser}
    \vspace{-3mm}
\end{figure}

Leveraging the strong image prior provided by SD, we propose the first diffusion-based zero-shot method for ODISR, named OmniSSR. Specifically, we propose Octadecaplex Tangent Information Interaction (OTII). OTII entails iterative conversion of intermediate SR results between ERP and TP representations, bridging the domain gap between ODIs and planar images. Building upon OTII, we further employ an approximate analytical solution of gradient descent, namely as Gradient Decomposition, to guide high-fidelity, high-quality omnidirectional image super-resolution. By capitalizing on SD's effective image prior, our approach strikes a balance between \textit{fidelity} and \textit{realness}, ensuring that the restored ODIs exhibit both fidelity to the input data and realistic visual details. This method shows potential for advancing the current state of ODISR, providing improved resolution and visual quality across various applications. Fig.~\ref{fig:teaser} showcases results fully demonstrating the superiority and performance of our proposed methods.

Our main contributions are summarized as follows:
\begin{itemize}
    \item We propose OmniSSR, the first zero-shot ODISR method, using an off-the-shelf diffusion-based model, requiring no training or fine-tuning, leveraging existing image generation model priors to solve ODISR task.
    \item To bridge the domain gap between ODIs and planar images, we introduce Octadecaplex Tangent Information Interaction by repeatedly transforming ODIs between ERP format and TP format, enabling ODISR task with pretrained diffusion models on planar images.
    \item By iteratively updating images using the developed Gradient Decomposition technique, we introduce consistency constraints into the sampling process of the latent diffusion model, ensuring a trade-off between fidelity and realness in the reconstructed results.
    \item Extensive experiments are conducted on the benchmark datasets, demonstrating the superior performance of our method over existing state-of-the-art approaches, which validate the effectiveness of OmniSSR.
\end{itemize}

\section{Related Work}

\subsection{Single Image Super-Resolution (SISR)}
Image super-resolution methods based on deep learning have undergone significant development over an extended period. Currently, they can be broadly classified into two categories of solutions. The first category involves end-to-end network training methods, which utilize image pairs consisting of low-resolution degraded images and high-resolution ground truth images for network training~\cite{mou2022mlim,Chen_2023_ICCV,Zhang_2021_ICCV,Lugmayr_2020_CVPR_Workshops,Lu_2022_CVPR,li2022d3c2,zhang2022herosnet, cheng2023hybrid}. The network architectures employed in this category include CNN~\cite{dong2015image}, Transformers~\cite{Vaswani_Shazeer_Parmar_Uszkoreit_Jones_Gomez_Kaiser_Polosukhin_2017}, etc. The second category employs image generation models as priors, such as GAN~\cite{gan}, diffusion models~\cite{ho2020denoising,song2020score}, etc., where low-resolution images are used as conditions to generate high-resolution images. We will mainly introduce the methods using generative prior.

\textbf{Single Image SR using GAN prior} 
In SR works utilizing GAN priors~\cite{menon2020pulse,daras2021intermediate,pan2021exploiting,chan2022glean,wang2023gpsr}, including real-world senarios~\cite{wang2018esrgan,wang2021real,mou2022mlim,zhang2024real}, pre-trained GAN networks are employed to transform image features into latent space, where the corresponding latent code for the high-resolution image is searched, ultimately yielding the reconstructed high-resolution result.

\textbf{Single Image SR using diffusion prior}
The diffusion model provides a powerful image prior, and the diffusion sampling process can generate highly realistic images. This strong prior distribution can be applied to various image restoration tasks, including super-resolution~\cite{wang2022zero,chung2022diffusion,chung2022improving,Song_Zhang_Yin_Mardani_Liu_Kautz_Chen_Vahdat,fei2023generative,sr3}. Image-domain diffusion models directly provide prior distributions of image-domain data. 
DDNM~\cite{wang2022zero} based on the mathematical method of Range-Nullspace Decomposition, iteratively refines content on the zero space, combining image prior content in the value domain to achieve image restoration. DDRM~\cite{kawar2022denoising} uses SVD decomposition to obtain restoration results, which is similar to DDNM. DPS~\cite{chung2022diffusion} transforms the image super-resolution problem into an optimization problem with consistency constraints, using gradient descent algorithms to guide the generation of image-domain diffusion models. GDP~\cite{fei2023generative} further uses such gradient to update the degradation operator to tackle blind image inverse problems. Other methods including MCG~\cite{chung2022improving}, DDS\cite{chung2022diffusion} and unified control of diffusion generation~\cite{Song_Zhang_Yin_Mardani_Liu_Kautz_Chen_Vahdat, fei2023generative} use same strategy for image restoration, especially image super-resolution.

The latent space diffusion model encodes data from various modalities into a latent space, samples its distribution, and then decodes it into the target domain. Image super-resolution works based on latent space domain include PSLD~\cite{Rout_Raoof_Daras_Caramanis_Dimakis_Shakkottai_2023}, P2L~\cite{Chung_Ye_Milanfar_Delbracio} and TextReg~\cite{kim2023regularization}. PSLD transfers the gradient-guided method of DPS~\cite{chung2022diffusion} to the latent space diffusion model, while P2L furthermore considers prompt design, iteratively optimizing the prompt embedding of SD to improve the quality and visual effects of image reconstruction. TextReg applies the textual description of the preconception of the solution of the image inverse problem during the reverse sampling phase, of which the description is dynamically reinforced through null-text optimization for adaptive negation.
\subsection{Omnidirectional Image Super-Resolution}
Omnidirectional image super-resolution (ODISR) aims to enhance the resolution of omnidirectional or 360-degree images, which are commonly captured by cameras with a wide field of view. This field has garnered increasing attention due to its applications in virtual reality, omnidirectional video streaming, and surveillance. Several approaches have been proposed to address the unique challenges of ODISR~\cite{arican2011joint,sun2023opdn,9506233,Cao_2023_CVPR,10222760}.
For instance, Kämäräinen et al.~\cite{Fakour-Sevom_Guldogan_Kamarainen_2018} propose a deep learning-based approach for omnidirectional super-resolution, leveraging convolutional neural networks to effectively upscale low-resolution omnidirectional images while preserving spatial details. Similarly, Smolic et al.~\cite{ozcinar2019super} introduce a novel omnidirectional super-resolution algorithm utilizing generative adversarial networks (GANs) to enhance the visual quality of omnidirectional images by hallucinating high-frequency details. 

For evaluation purposes, researchers commonly utilize datasets such as the ODI-SR dataset from LAU-Net~\cite{Deng_Wang_Xu_Guo_Song_Yang_2021}, and SUN 360 Panorama dataset~\cite{Jianxiong_Xiao_Ehinger_Oliva_Torralba_2012}. These datasets enable the quantitative assessment of ODISR algorithms across various scenarios and facilitate fair comparisons between different methods.

\section{Method}
In this section, we first briefly introduce the preliminary background of our method (Sec.~\ref{subsec:preliminaries}), and give an overall view of our proposed OmniSSR (Sec.~\ref{subsec:overview}). Then, we discuss the designs of Octadecaplex Tangent Information Interaction, which transform ODIs between ERP and TP formats with pre-upsampling strategy (Sec~\ref{subsec:transform}), and the Gradient Decomposition correction (Sec.~\ref{subsec:rnd}).
\subsection{Preliminaries}
\label{subsec:preliminaries}
\subsubsection{ERP$\leftrightarrow$TP Transformation}
\label{subsec:e2t}
The essence of projection transformations between ERP and TP lie in determining the positions of target image pixels within the source image and computing their corresponding pixel values using interpolation algorithms, as digital images are always stored discretely~\cite{li2022omnifusion}. Therefore, the ERP$\rightarrow$TP transformation involves locating the TP image pixels on the ERP imaging plane, and vice versa. Gnomonic projection~\cite{coxeter1961introduction} provides the correspondence between ERP image pixels and TP image pixels.


For a pixel $P_e(x_e, y_e)$ within the ERP image, we first find its corresponding pixel $P_s(\theta, \phi)$ on the unit sphere using Eq.~\ref{eq:erp2sphere}:
\begin{equation}
    \theta = 2\pi x_e / W,\; \phi = \pi y_e / H,
\label{eq:erp2sphere}
\end{equation}
where $H$ and $W$ are the height and width of the ERP image.
The Cartesian coordinates of the ERP image and the angular coordinates on the unit sphere exhibit a straightforward one-to-one linear relationship, suggesting a conceptual equivalence between these two projection formats.

Given the spherical coordinates of the tangent plane center $(\theta_c, \phi_c)$, The transformation from $P_s(\theta, \phi)$ to $P_t(x_t, y_t)$, i.e. ERP$\rightarrow$TP, is defined as:
\begin{equation}
\begin{aligned}
    &x_t = \big( \cos(\phi)\sin(\theta - \theta_c) \big) \big/ \zeta , \\
    &y_t = \big( \cos(\phi_c)\sin(\phi) - \sin(\phi_c)\cos(\phi)\cos(\theta - \theta_c) \big) \big/ \zeta ,
\end{aligned}
\label{eq:sphere2tan}
\end{equation}
where $\zeta = \sin(\phi_c)\sin(\phi) + \cos(\phi_c) \cos(\phi) \cos(\theta - \theta_c)$.

The corresponding inverse transformation, i.e. TP$\rightarrow$ERP, is:
\begin{equation}
\begin{aligned}
    &\theta = \theta_c + \arctan \big( (x_t\sin(c)) / (\rho\cos(\phi_1)\cos(c) - y_t\sin(\phi_c)\sin(c)) \big), \\
    &\phi = \arcsin \big( \cos(c)\sin(\phi_c) +  y_t  \sin(c)\cos(\phi_c) / \rho \big),
\end{aligned}
\label{eq:tan2sphere}
\end{equation}
where $\rho = \sqrt{x_t^2 + y_t^2}$ and $c=\arctan(\rho)$.

\begin{figure}
    \centering
    \includegraphics[width=1\linewidth]{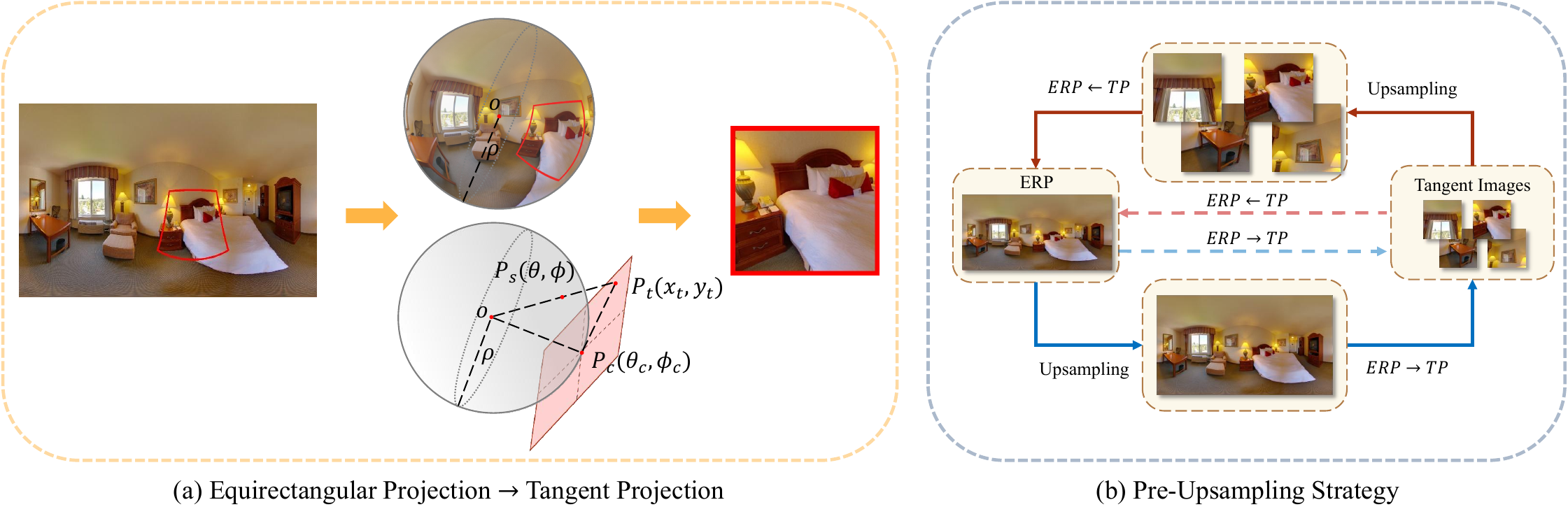}
    \caption{Details about gnomonic transformations. (a) conversion from ERP to TP. (b) pre-upsampling proposed in Octadecaplex Tangent Information Interaction (Sec.~\ref{subsec:transform}) mitigating loss during transformation.}
    \label{fig:otp}
    \vspace{-4mm}
\end{figure}

With Eq.~\ref{eq:sphere2tan} and Eq.~\ref{eq:tan2sphere}, we can build one-to-one forward and inverse mapping functions between pixels on the ERP image and pixels on the TP images. An illustration of the ERP$\rightarrow$TP transformation is shown in Fig.~\ref{fig:otp}(a).

\subsubsection{Iterative Denoising for Super-Resolution}
\label{subsec:sdsr}
Utilizing the rich image priors provided by SD, we can super-resolve planar images. During initialization, the images are passed through the encoder $\mathcal{E}$ of SD to obtain latent codes, which are then added to pure noise to generate initial noise $\mathbf{z}_{T}$. Following the approach proposed by StableSR~\cite{StableSR_Wang_Yue_Zhou_Chan_Loy_2023}, we pass the images through a time-aware adapter $\mathcal{T}$. This adapter network structure is similar to the down-sampling part in denoising UNet, taking the image and the time step $t$ of diffusion sampling as inputs to obtain the latent code feature for step $t$. This feature, along with the latent code $\mathbf{z}_{t}$ for each step and the time step $t$, is then passed through denoising UNet to calculate the denoised result $\mathbf{z}_{0|t}$ and the latent code $\mathbf{z}_{t-1}$ for the next sampling step. 
By iterating this process $T$ times, we can obtain the final super-resolution result via decoder $\mathcal{D}$ of SD, yielding high-resolution images.
\subsection{Overview}
\label{subsec:overview}
\begin{figure}
    \centering
    \vspace{-3mm}
    \includegraphics[width=1\linewidth]{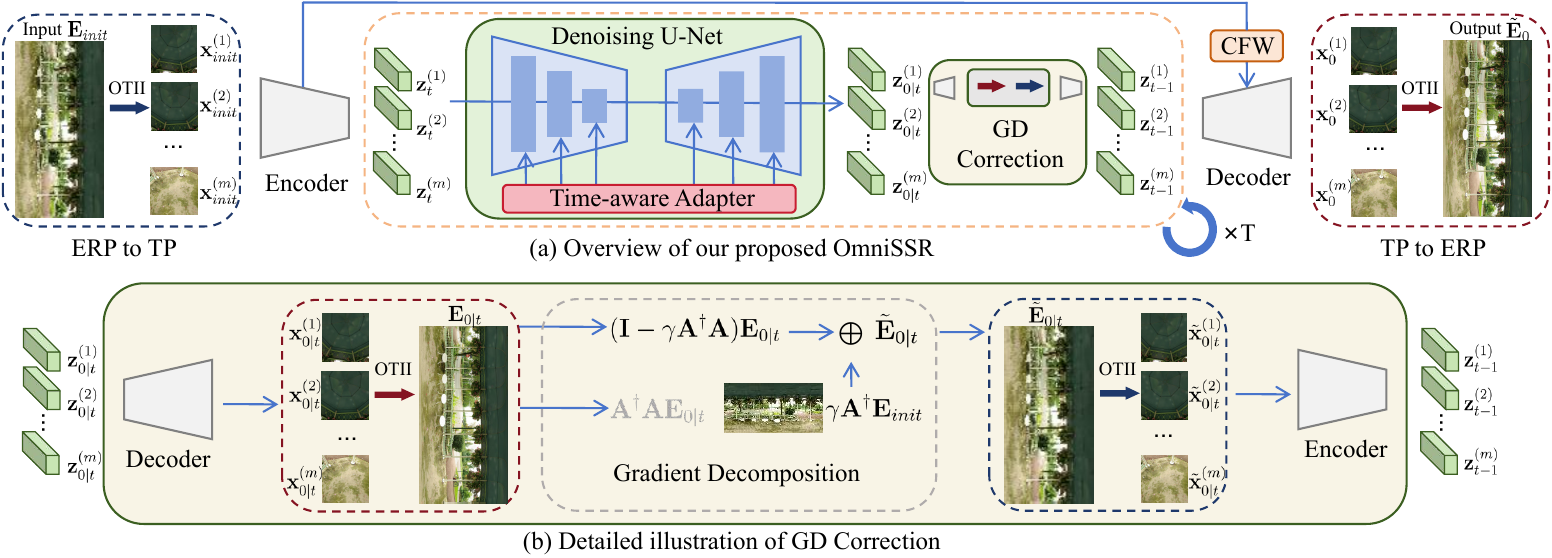}
    \caption{Overview of our proposed OmniSSR. Input low-resolution omnidirectional image $\mathbf{E}_{init}$ in ERP format is first projected onto Tangent Projection (TP) images $\mathbf{x}_{init}^{(1)},\mathbf{x}_{init}^{(2)},...,\mathbf{x}_{init}^{(m)}$, then iteratively refined via Stable Diffusion (SD) with a time-aware adapter and controllable feature wrapping (CFW) module. In each step of diffusion sampling, we adopt the Gradient Decomposition (GD) correction technique to introduce consistency constraints for the restored intermediate results. After $T$ steps of sampling, we obtain the final result $\mathbf{\tilde{\mathbf{E}}}_{0}$ with high resolution and better visual quality.}
    \vspace{-5mm}
    \label{fig:overview}
\end{figure}
Our approach can be divided into three parts. The first part is pre-processing, where we initially up-sample the low-resolution ERP images $\mathbf{E}_{init}$, then project them onto tangent planes to obtain a series of TP images. These TP images are transformed to the latent space by the SD encoder, iteratively processed through denoising UNet and time-aware adapter network, and then decoded to obtain high-resolution TP images. During each denoising step, these TP images are transformed back via inverse transformation to ERP images, employing the Gradient Decomposition correction to ensure consistency constraints in diffusion sampling. After $T$ iterations, the final super-resolution result is obtained. A formulaic description for OmniSSR pipeline is shown in Algo.~\ref{alg:pipeline}.
Fig.~\ref{fig:overview} shows the overview of our proposed pipeline.
\begin{figure}[t]
 \begin{minipage}[t]{.45\textwidth}
    \begin{algorithm}[H]
    \scriptsize
    \caption{OmniSSR Pipeline}
    \label{alg:pipeline}
    \KwIn{$\mathbf{E}_{init}$, $\mathcal{F}$, $\mathcal{F}^{-1}$, $\mathbf{A}$, $\mathbf{A}^{\dagger}$, $\mathcal{E}$, $\mathcal{D}$, $T$}
    \KwOut{SR result $\tilde{\mathbf{E}}_{0}$}
    $\{ \mathbf{x}_{init}^{(1)},\mathbf{x}_{init}^{(2)},...,\mathbf{x}_{init}^{(m)} \} = \mathcal{F}(\mathbf{E}_{init})$ \\
    \For{$i=1$ \KwTo $m$}{
        $\mathbf{z}_{init}^{(i)} = \mathcal{E}(\mathbf{x}_{init}^{(i)})$\\
        $\boldsymbol{\epsilon}^{(i)} \sim \mathcal{N}(\mathbf{0},\mathbf{I})$\\
        $ \mathbf{z}_{T}^{(i)}=\sqrt{\overline{\alpha}_{T}}\mathbf{z}_{init}^{(i)} + \sqrt{1-\overline{\alpha}_T}\boldsymbol{\epsilon}^{(i)} $\\
    }
    Get$\{ \mathbf{z}_{0}^{(1)},\mathbf{z}_{0}^{(2)},...,\mathbf{z}_{0}^{(m)} \}$ from  Algo.~\ref{alg:sdsr+rnd} \\
    \For{$i=1$ \KwTo $m$}{
        $ \mathbf{x}_{0}^{(i)} = \mathcal{D}(\mathbf{z}_{0}^{(i)}) $
    }
    
    $\mathbf{E}_{0}=\mathcal{F}^{-1}(\{ \mathbf{x}_{0}^{(1)},
    \mathbf{x}_{0}^{(2)} ,...,
    \mathbf{x}_{0}^{(m)} \})$ \\
    $\tilde{\mathbf{E}}_{0}= \mathbf{E}_{0} + \gamma_{p}\mathbf{A}^{\dagger}(\mathbf{E}_{init}-\mathbf{A}\mathbf{E}_{0})$ \\ \label{line:gd-post}
    \Return $\tilde{\mathbf{E}}_{0}$
    \end{algorithm}
\end{minipage}
\begin{minipage}[t]{.51\textwidth}
  \centering
  \begin{algorithm}[H]
    \scriptsize
    \caption{Iterative Denoising with GD Correction}
    \label{alg:sdsr+rnd}
    \KwIn{$\mathbf{E}_{init}$, $\mathcal{F}$, $\mathcal{F}^{-1}$, $\mathbf{A}$, $\mathbf{A}^{\dagger}$, $\mathcal{E}$, $\mathcal{D}$, $\mathcal{T}$, $T$}
    \KwOut{Latent code $\{ \mathbf{z}_{0}^{(1)},\mathbf{z}_{0}^{(2)},...,\mathbf{z}_{0}^{(m)} \}$}
    \For{$t=T$ \KwTo $1$}{
    \For{$i=1$ \KwTo $m$}{
    $\boldsymbol{\epsilon}_{t}=\boldsymbol{\epsilon}_{\theta}(\mathbf{z}_{t}^{(i)},\mathcal{T}(\mathbf{z}_{init}^{(i)},t),t)$\\
    $\mathbf{z}_{0|t}^{(i)}=\frac{1}{\sqrt{\overline{\alpha}_{t}}}(\mathbf{z}_{t}^{(i)}-\boldsymbol{\epsilon}_{t}\sqrt{1-\overline{\alpha}_{t}})$ \\ 
    $\mathbf{x}_{0|t}^{(i)}=\mathcal{D}(\mathbf{z}_{0|t}^{(i)})$\\
    }
    $\mathbf{E}_{0|t}=\mathcal{F}^{-1}(\{ \mathbf{x}_{0|t}^{(1)},\mathbf{x}_{0|t}^{(2)},...,\mathbf{x}_{0|t}^{(m)} \})$\\ \label{line:otii}
    $\tilde{\mathbf{E}}_{0|t}=\mathbf{E}_{0|t} + \gamma_{e}\mathbf{A}^{\dagger}(\mathbf{E}_{init}-\mathbf{A}\mathbf{E}_{0|t})$ \\ \label{line:gammae}
    $\{ \tilde{\mathbf{x}}_{0|t}^{(1)},\tilde{\mathbf{x}}_{0|t}^{(2)},...,\tilde{\mathbf{x}}_{0|t}^{(m)} \}=\mathcal{F}(\tilde{\mathbf{E}}_{0|t})$\\
    \For{$i=1$ \KwTo $m$}{
    $\tilde{\mathbf{z}}_{0|t}^{(i)}= (1-\gamma_{l}) \mathbf{z}_{0|t}^{(i)} + \gamma_{l} \mathcal{E}(\tilde{\mathbf{x}}_{0|t}^{(i)})$\\ \label{line:rnd-latent}
    $\mathbf{z}_{t-1}^{(i)} \sim p(\mathbf{z}_{t-1}^{(i)}|\mathbf{z}_{t}^{(i)},\tilde{\mathbf{z}}_{0|t}^{(i)})$\\
        }
    }
    \Return $\{ \mathbf{z}_{0}^{(1)},\mathbf{z}_{0}^{(2)},...,\mathbf{z}_{0}^{(m)} \}$
    \end{algorithm}
  \end{minipage}
\vspace{-3mm}
\end{figure}


\subsection{Octadecaplex Tangent Information Interaction (OTII)}
\label{subsec:transform}

\subsubsection{Motivation}
To apply SD for ODISR, a straightforward way is to perform the ERP$\rightarrow$TP transformation on the input ERP image. Then, each obtained TP image is fed into the SD-based model for SR. Finally, the TP$\rightarrow$ERP transformation yields the ultimate super-resolved ERP image. OmniFusion~\cite{li2022omnifusion} employs a similar approach for depth estimation. However, this simplistic strategy fractures the inherent global coherence of ODIs, leading to pixel-level discontinuities in the fused ERP images. Moreover, interpolation algorithms cause significant information loss in the original projection transformations, resulting in more blurred images. If applied multiple times, this exacerbates the information loss even further. To mitigate this, a trivial solution is to increase the pixel count (resolution) of the intermediate projection imaging plane. However, excessively high resolutions in TP images can introduce unnecessary computational overhead during the denoising stage and potentially compromise the denoising performance.

\subsubsection{Information Interaction and Pre-upsampling}

Based on the observations and analysis presented above, we propose OTII by alternating the intermediate results between ERP and TP formats at each denoising step, where a single ERP image is represented by 18 TP images. From Sec.~\ref{subsec:e2t}, we can achieve the ERP$\rightarrow$TP transformation (denoted as $\mathcal{F}(\cdot)$) and the TP$\rightarrow$ERP transformation (denoted as $\mathcal{F}^{-1}(\cdot)$). Through the ERP$\rightarrow$TP transformation, we can convert distorted ERP images into TP images with content distributions that approximate those of planar images. This enables the use of the original SD super-resolution method for planar images. Conversely, through the TP$\rightarrow$ERP transformation, we can fuse information between different TP images holistically, while providing ERP-format input for the subsequent GD Correction in Sec.~\ref{subsec:rnd}. To handle information loss during projection transformation, we further propose to pre-upsample the source image before projection transformations, as shown in Fig.~\ref{fig:otp}(b). Our experiments in Sec.~\ref{sebsec: ablation study} demonstrate that this pre-upsampling strategy can significantly mitigate the information loss caused by projection transformations.

\subsection{Gradient Decomposition (GD) Correction for Fidelity}
\label{subsec:rnd}
SD-based methods, as introduced in Sec.~\ref{subsec:sdsr}, can perform SR on sliced TP images. However, relying solely on the SR results from SD may lack consistency and fail to accurately preserve the original semantic information and details of the low-resolution image.\footnote{This claim will be further illustrated in subsequent experiments.}
To enhance the consistency of the SR results from SD, we opt to use convex optimization methods to iteratively refine them. Modeling the SR task as an image inverse problem, the following equation is formulated:
\begin{equation}
    \mathbf{y}=\mathbf{Ax}+\mathbf{n}, \quad \mathbf{n} \sim \mathcal{N}(\mathbf{0},\mathbf{I}),
\end{equation}
where $\mathbf{x}$ represents the original image, $\mathbf{y}$ denotes the degraded result, $\mathbf{A}$ is the degradation operator (e.g., bicubic downsampling for super-resolution), and $\mathbf{n}$ is random noise. The objective we aim to solve can be expressed as the following convex optimization problem:
\begin{equation}
    \underset{\mathbf{x}}{\mathrm{argmin}}
    ||\mathbf{y}-\mathbf{Ax}||_{2}^{2} + \lambda \mathcal{R}(\mathbf{x}),
\end{equation}
where the first term is the data-fidelity term, ensuring the consistency of image reconstruction, and the second term is the regulation term, ensuring the sparsity of the reconstruction result, thus making the image more realistic. The regularization term can be the 1-norm, Total Variation, etc.
The aforementioned convex optimization problem can be solved using a series of algorithms, such as gradient descent, ADMM, etc. Considering the trade-off between time and performance, we turn to find a solution based on gradient descent, and provide an approximate analytical solution composed of a \textit{fidelity} term and a \textit{realness} term, named "Gradient Decomposition (GD)":
\begin{equation}
\begin{aligned}
    \tilde{\mathbf{E}}_{0|t}
    & =\mathbf{E}_{0|t}+\alpha \nabla_{\mathbf{E}_{0|t}}||\mathbf{E}_{init}-\mathbf{A}\mathbf{E}_{0|t}||_{F} 
    = \mathbf{E}_{0|t}+\alpha \times 2   (\mathbf{A}^{\dagger}\mathbf{E}_{init}-\mathbf{A}^{\dagger}\mathbf{A}\mathbf{E}_{0|t}) \\
    & = \mathbf{E}_{0|t}+\gamma \mathbf{A}^{\dagger}(\mathbf{E}_{init}-\mathbf{A}\mathbf{E}_{0|t}) 
    = \gamma \mathbf{A}^{\dagger}\mathbf{E}_{init} + (\mathbf{I}-\gamma \mathbf{A}^{\dagger}\mathbf{A})\mathbf{E}_{0|t}
\end{aligned}
\label{eq:gdc}
\end{equation}
where $\mathbf{A}^{\dagger}$denotes pseudo-inverse of degradation operator $\mathbf{A}$, $\mathbf{E}_{init}$ denotes initial low-resolution ERP input, $\mathbf{E}_{0|t}$ denotes restored result by SD, $\tilde{\mathbf{E}}_{0|t}$ denotes corrected result by GD, $\alpha$ denotes the learning rate of gradient descent, and $\gamma$ denotes the simplified hyper-parameter which is further tuned using grid search. The final setting of $\gamma$ on different stages is shown in Sec.~\ref{sec:settings}, and the ablation studies of parameter choice are in Sec.~\ref{sebsec: ablation study}.

This technique could be seen as a step of gradient descent optimization, and the optimized result could be decomposed of (1) $\gamma \mathbf{A}^{\dagger}\mathbf{E}_{init}$, which ensures the consistency of the generated result, and (2) $(\mathbf{I}-\gamma \mathbf{A}^{\dagger}\mathbf{A})\mathbf{E}_{0|t}$, which serves as the iteratively updated generated result to improve its realness; $\gamma$ is a hyper-parameter balancing the data fidelity and visual quality. For a better diversity and generality of the SR process, we expand this solution to latent space, and obtain the denoising result from both denoising UNet and corrected TP images (Algo.~\ref{alg:sdsr+rnd} line~\ref{line:rnd-latent}). A more detailed understanding of the iterative denoising process and application of GD correction could be referred to Algo.~\ref{alg:sdsr+rnd}.

\section{Experiments}
\label{sec:experiments}
\subsection{Implementation Details}
\subsubsection{Datasets and Pretrained Models}
We choose the test set of ODI-SR dataset from LAU-Net~\cite{Deng_Wang_Xu_Guo_Song_Yang_2021} and SUN 360 Panorama dataset~\cite{Jianxiong_Xiao_Ehinger_Oliva_Torralba_2012}, comprising 97 and 100 omnidirectional images respectively, for experimental evaluation. The ground truth images are of size 1024$\times$2048 pixels. In SR methods such as GDP~\cite{fei2023generative} and PSLD~\cite{Rout_Raoof_Daras_Caramanis_Dimakis_Shakkottai_2023} for planar images, we partitioned the images into several 256$\times$256 patches and performed super-resolution on each patch individually.

For pre-trained models, we adopt from StableSR~\cite{StableSR_Wang_Yue_Zhou_Chan_Loy_2023}, which provided a SR network for planar images based on SD. This network architecture includes a time-aware adapter, a controllable feature wrapping (CFW) module, and the original SD structure from HuggingFace. All of them are kept untrained in our proposed OmniSSR.

\subsubsection{Settings}
\label{sec:settings}
We set diffusion sampling steps to 200, which is the same as StableSR. The steps for other diffusion-based methods are set the same as their default settings (e.g. 1000 steps for PSLD).
The degradation for low-resolution ERP images is bicubic down-sampling, and the implementation of its pseudo-inverse can be referred from code of DDRM~\cite{kawar2022denoising}\footnote{https://github.com/bahjat-kawar/ddrm}. For choices of hyper-parameter $\gamma$ in GD correction, we set $\gamma_{p}=1.0$, $\gamma_{e}=1.0$, $\gamma_{l}=0.5$. Our code is developed via PyTorch on NVIDIA 3090Ti GPU. 
\footnote{Code will be made available.}
\begin{table}[h] 
	\centering
	\scriptsize
    \caption{SR results under bicubic downsampling on ODI-SR and SUN 360 Panorama datasets. For tasks not implemented in those papers, we mark N/A in corresponding results. Best results are shown in \textcolor{Red}{\textbf{Red}}, and second best results are shown in \textcolor{Blue}{\textbf{Blue}}.}
		\resizebox{1.0\linewidth}{!}{
		\begin{tabular}{c|c|cccc|cccc}
			\toprule
			\multicolumn{1}{c|}{\multirow{2}{*}{Method}}  
            & \multirow{2}{*}{Scale}        
            & \multicolumn{4}{c|}{ODI-SR}         & \multicolumn{4}{c}{SUN 360 Panorama} \\
			\multicolumn{1}{c|}{}&&                                     WS-PSNR$\uparrow$ & WS-SSIM$\uparrow$ & FID$\downarrow$ & LPIPS$\downarrow$  & WS-PSNR$\uparrow$ & WS-SSIM$\uparrow$ & FID$\downarrow$ & LPIPS$\downarrow$  \\ \midrule
			Bicubic   
            & \multirow{8}{*}{$\times$2}
            &28.14&0.8343&24.00&0.2164
            &28.67&0.8537&29.25&0.1933   \\ 
			DDRM~\cite{kawar2022denoising}     
            &&\boldblue{27.90}&\boldblue{0.8317}&\boldred{12.28}&\boldblue{0.1661}
            &\boldblue{29.55}&\boldblue{0.8670}&\boldblue{13.10}&\boldred{0.1426}\\
            DPS~\cite{chung2022diffusion}    
            &&20.99&0.6194&148.30&0.5249 
            &21.44&0.6598&148.83&0.5175\\
           GDP~\cite{fei2023generative}   
           &&27.89&0.8157&26.56&0.2724
           &28.60&0.8376&28.02&0.2445\\
			PSLD~\cite{Rout_Raoof_Daras_Caramanis_Dimakis_Shakkottai_2023}                               
            && N/A   & N/A    & N/A   & N/A  
            & N/A   & N/A    & N/A   & N/A  \\ 
            DiffIR~\cite{Xia_Zhang_Wang_Wang_Wu_Tian_Yang_Gool}   
            && 23.77& 0.6583&57.23&0.4687
            &23.54&0.6775&58.06&0.4658\\
		  StableSR~\cite{StableSR_Wang_Yue_Zhou_Chan_Loy_2023}           
            && 22.70   & 0.6458    & 44.87   & 0.3039  
            & 23.30   & 0.6907    & 43.49   & 0.2858   \\
		  OmniSSR 
            &&\boldred{28.57}&\boldred{0.8540}& \boldblue{13.01}   &\boldred{0.1575}
            & \boldred{29.69}   & \boldred{0.8781}    & \boldred{12.99}   & \boldblue{0.1459}   \\ 
			 
            \hline 
			Bicubic 
            & \multirow{8}{*}{$\times$4}
            &25.43&0.7059&50.84&0.3755
            & 25.49   & 0.7229    & 55.99   & 0.3656   \\ 
			DDRM~\cite{kawar2022denoising}        
            && \boldblue{25.43}&\boldred{0.7367}&\boldblue{32.69}&0.3206
            & \boldblue{25.83}   & \boldblue{0.7443}    & \boldred{32.93}   & 0.3304 \\
            DPS~\cite{chung2022diffusion}      
            && 24.75   & 0.6594    & 120.74   & 0.4911 
            & 21.09   & 0.6119    & 175.2143   & 0.5541  \\
            GDP~\cite{fei2023generative}   	                  
            && 23.16   & 0.6692	   & 77.43   &0.4260
            & 23.75      & 0.6569    & 90.23   & 0.4240  \\
			PSLD~\cite{Rout_Raoof_Daras_Caramanis_Dimakis_Shakkottai_2023}
            && 21.72   & 0.5498    & 107.99   & 0.5329  
            & 21.75   & 0.5828    & 141.49   & 0.5461  \\ 
            DiffIR~\cite{Xia_Zhang_Wang_Wang_Wu_Tian_Yang_Gool}                               
            && 24.01   & 0.6770    &54.14&0.4367
            & 23.90   & 0.7014    & 50.37   & 0.4235  \\
			StableSR~\cite{StableSR_Wang_Yue_Zhou_Chan_Loy_2023}
            && 23.33   & 0.6577    & 49.95   & \boldblue{0.3135} 
            & 23.99 & 0.6998    & 46.03   & \boldblue{0.3023}   \\
			OmniSSR 
            &&\boldred{25.77}&\boldblue{0.7279}& \boldred{30.97}   &\boldred{0.2977}
            & \boldred{26.01}   &\boldred{0.7481}    & \boldblue{34.58}   & \boldred{0.2963}   \\ \bottomrule
		\end{tabular}
		}
	
	\label{tab:main}
	\vspace{-3mm}
\end{table}

\subsection{Comparison of OmniSSR with diffusion-based methods}
\label{sec:eval}
To evaluate the performance of proposed OmniSSR, we compare our method with recent state-of-the-art zero-shot methods for single image SR task: 
DPS~\cite{chung2022diffusion},
DDRM~\cite{kawar2022denoising},
GDP~\cite{fei2023generative} which are based on the image-domain diffusion model, and PSLD~\cite{Rout_Raoof_Daras_Caramanis_Dimakis_Shakkottai_2023}, which is based on latent diffusion model. We also choose supervised diffusion-based super-resolution approaches including StableSR~\cite{StableSR_Wang_Yue_Zhou_Chan_Loy_2023} and DiffIR~\cite{Xia_Zhang_Wang_Wang_Wu_Tian_Yang_Gool} for comparison. We conduct experiments on $\times$2 and $\times$4 SR with ERP bicubic downsampling, on ODI-SR test-set and SUN test-set. 
We choose WS-PSNR~\cite{WS_PSNR_Sun_Lu_Yu_2017}, WS-SSIM~\cite{WS_SSIM_Zhou_Yu_Ma_Shao_Jiang_2018}, FID~\cite{fid}, and LPIPS~\cite{zhang2018perceptual} as the main metrics. 

\textbf{Quantitative results} are presented in 
Tab.~\ref{tab:main}. With proposed OTII and GD correction, OmniSSR out-performs previous methods in terms of both \textit{Fidelity} (from WS-PSNR and WS-SSIM) and \textit{Realness} (from FID, LPIPS ), which shows superior performance to existing diffusion-based methods for ODISR tasks on different scales.

\begin{figure*}[h]
	\scriptsize
	\centering
	\begin{tabular}{l}
		\hspace{-0.42cm}
		\begin{adjustbox}{valign=t}
			\begin{tabular}{c}
				\includegraphics[width=0.260\textwidth]{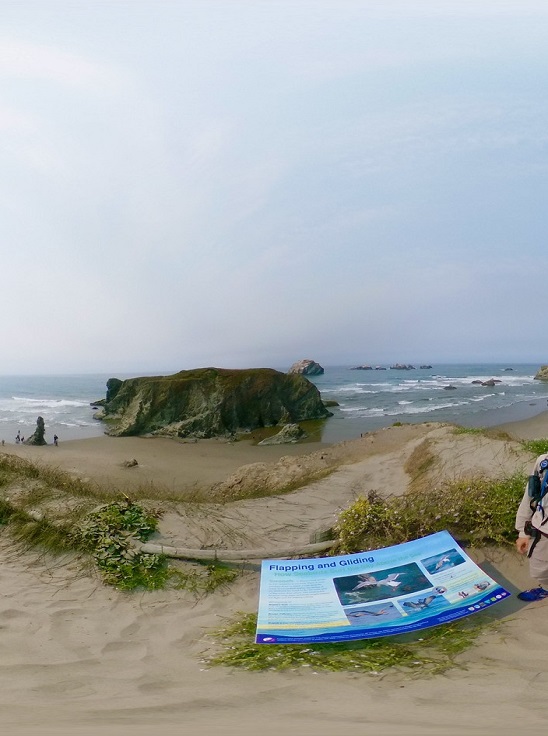}
				\vspace{1.5mm}
				\\
				SUN 360 ($\times$2): 001
			\end{tabular}
		\end{adjustbox}
		\hspace{-2mm}
		\begin{adjustbox}{valign=t}
			\begin{tabular}{cccc}
				\includegraphics[width=0.149\textwidth]{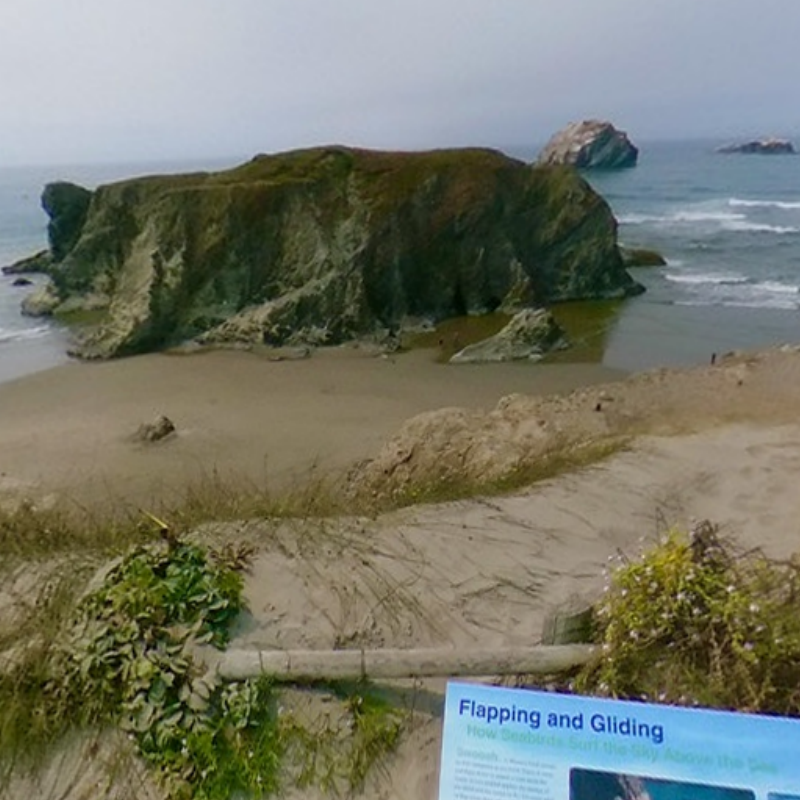} \hspace{-1mm} &
				\includegraphics[width=0.149\textwidth]{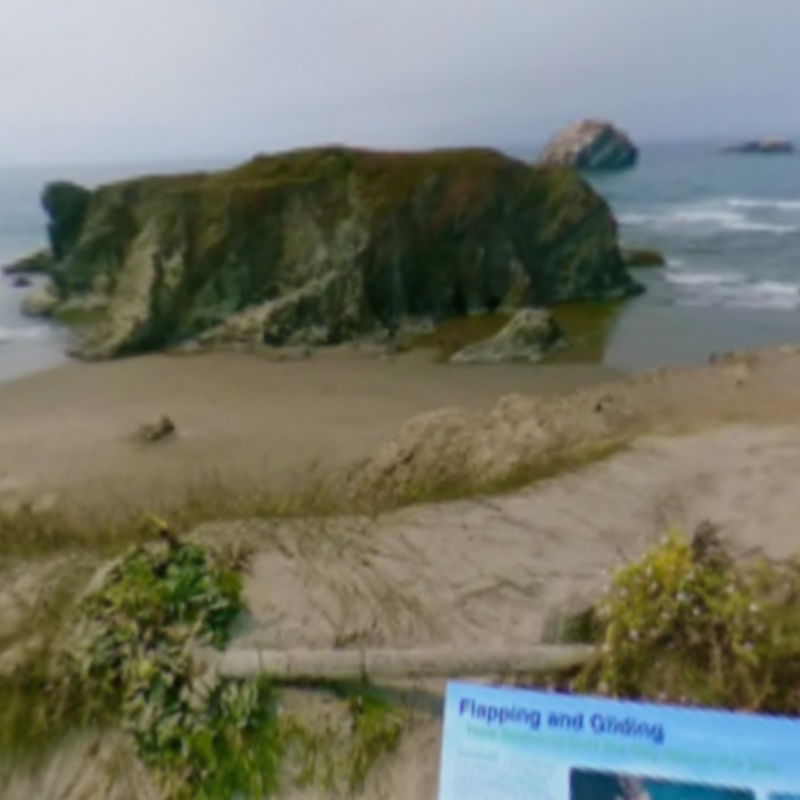} \hspace{-1mm} &
				\includegraphics[width=0.149\textwidth]{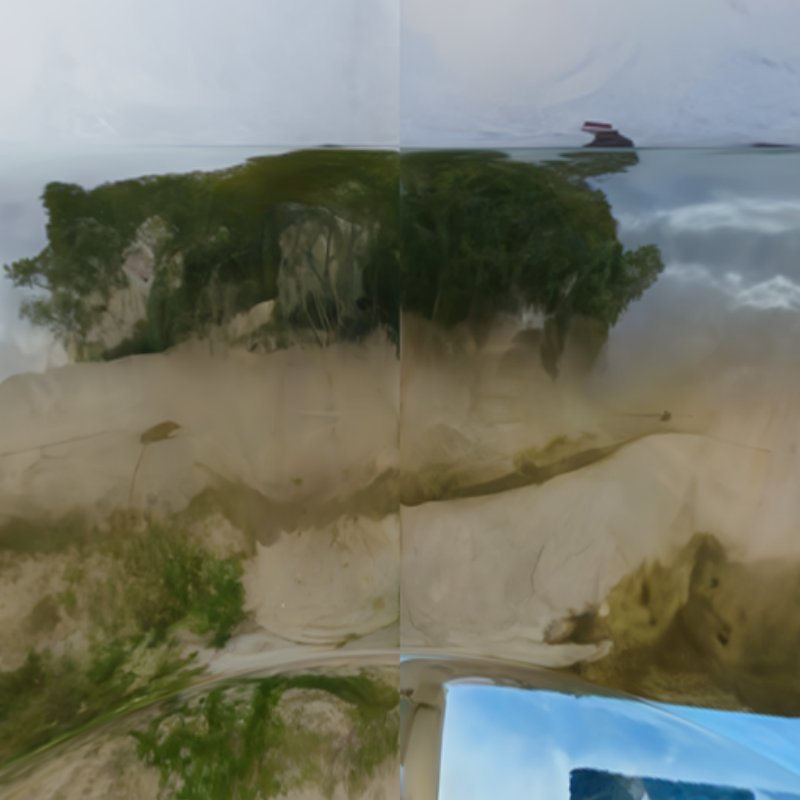} \hspace{-1mm} &
				\includegraphics[width=0.149\textwidth]{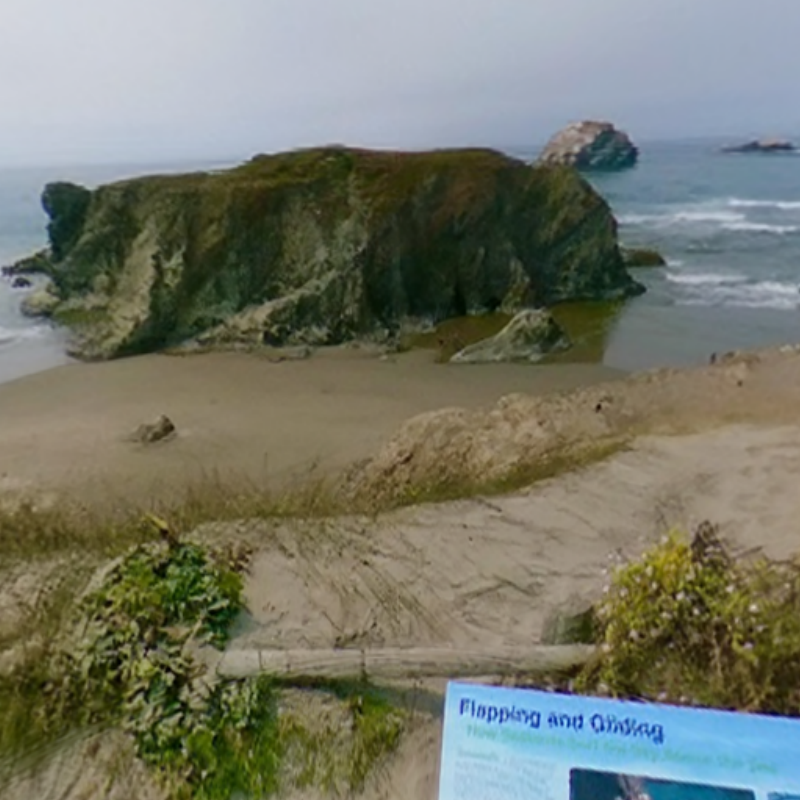} \hspace{-1mm} 
				\\
				HR \hspace{-1mm} &
				Bicubic \hspace{-1mm} &
				DPS \cite{chung2022diffusion} \hspace{-1mm} &
				DDRM \cite{kawar2022denoising} \hspace{-1mm} 
				\\ 
				PSNR/SSIM \hspace{-1mm} &
				28.67dB/0.8317 \hspace{-1mm} &
				24.02dB/0.5849 \hspace{-1mm} &
				30.04dB/0.8855 \hspace{-1mm} 
				\\
				\includegraphics[width=0.149\textwidth]{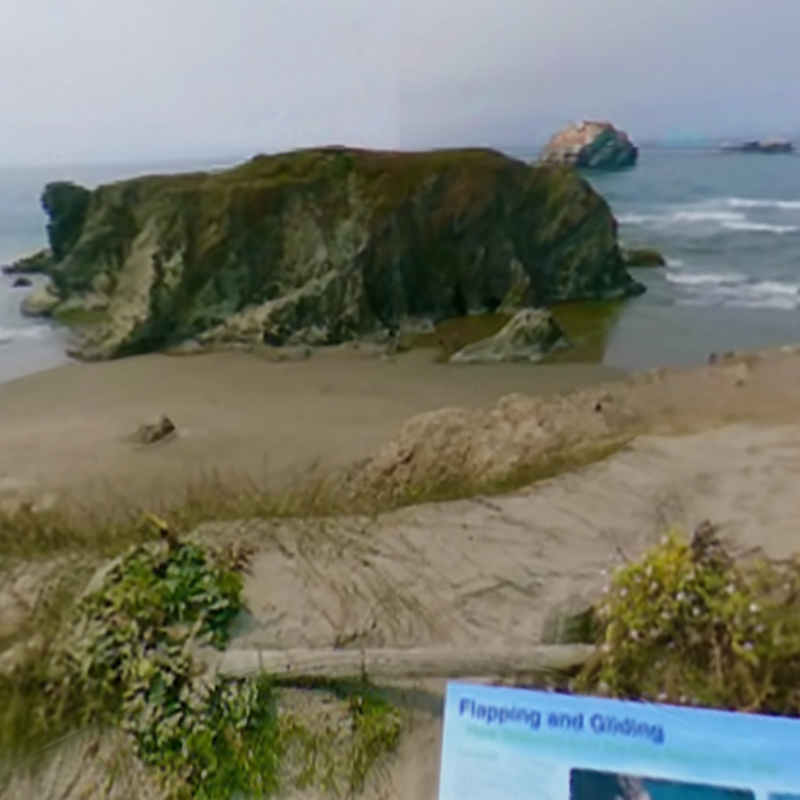} \hspace{-1mm} &
				\includegraphics[width=0.149\textwidth]{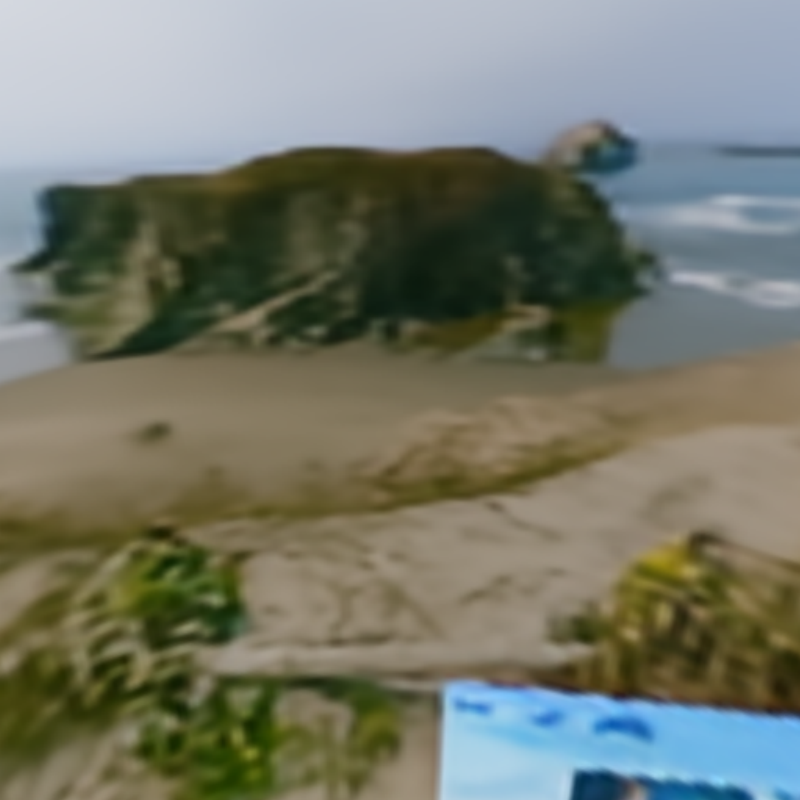} \hspace{-1mm} &
				\includegraphics[width=0.149\textwidth]{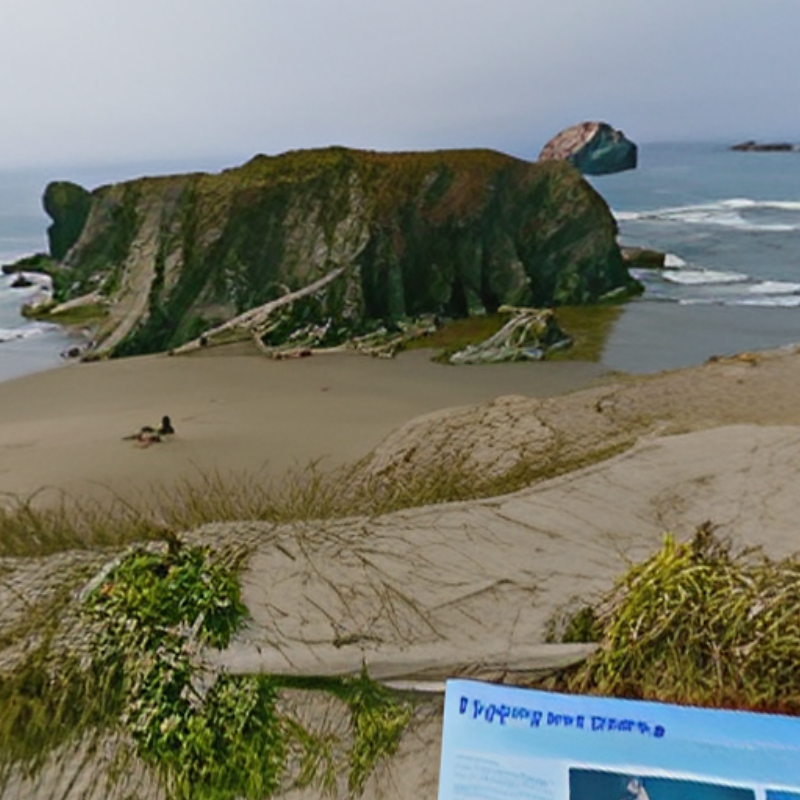} \hspace{-1mm} &
				\includegraphics[width=0.149\textwidth]{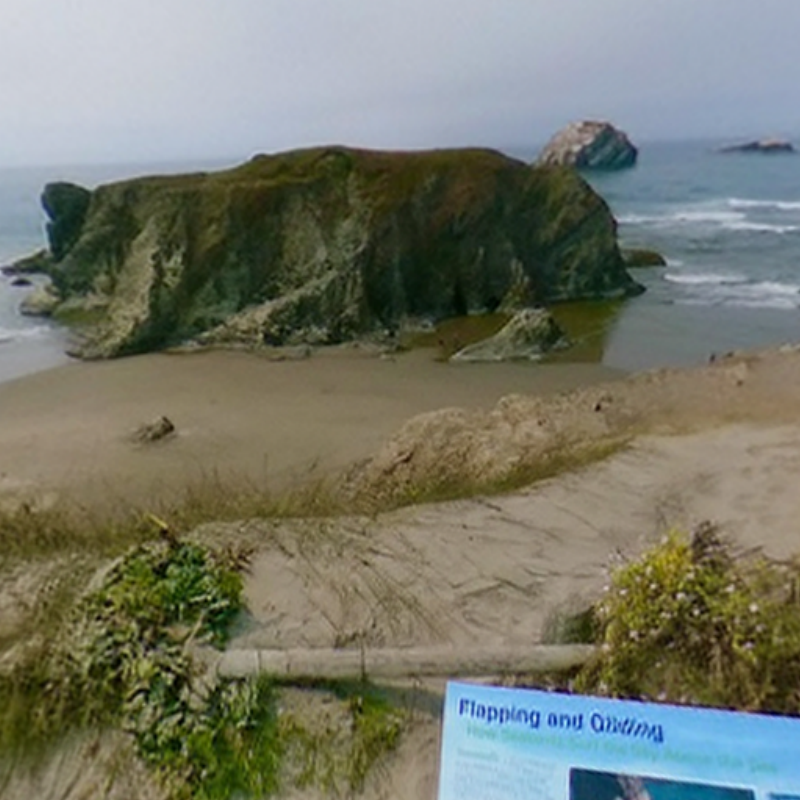} \hspace{-1mm}  
				\\ 
				GDP \cite{fei2023generative} \hspace{-1mm} &
				DiffIR \cite{Xia_Zhang_Wang_Wang_Wu_Tian_Yang_Gool} \hspace{-1mm} &
				StableSR \cite{StableSR_Wang_Yue_Zhou_Chan_Loy_2023} \hspace{-1mm} &
				\textbf{OmniSSR}(ours) \hspace{-1mm} 
				\\
				30.07dB/0.8802 \hspace{-1mm} &
				24.47dB/0.6421 \hspace{-1mm} &
				23.81dB/0.7384 \hspace{-1mm} &
				30.15dB/0.8859 \hspace{-1mm} 
			\end{tabular}
		\end{adjustbox}
		\vspace{2mm}
		
		\\ 
		\hspace{-0.42cm}
		\begin{adjustbox}{valign=t}
			\begin{tabular}{c}
				\includegraphics[width=0.260\textwidth]{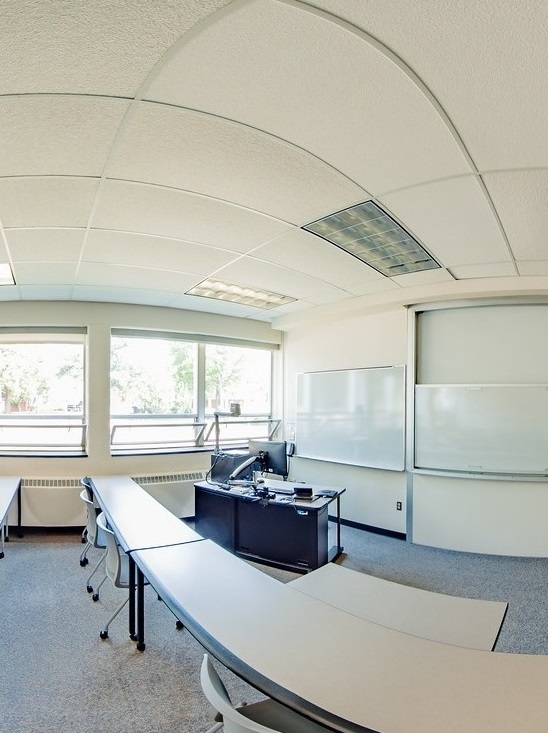}
				\vspace{1.5mm}
				\\
				SUN 360 ($\times$4): 009
			\end{tabular}
		\end{adjustbox}
		\hspace{-2mm}
		\begin{adjustbox}{valign=t}
			\begin{tabular}{cccc}
				\includegraphics[width=0.149\textwidth]{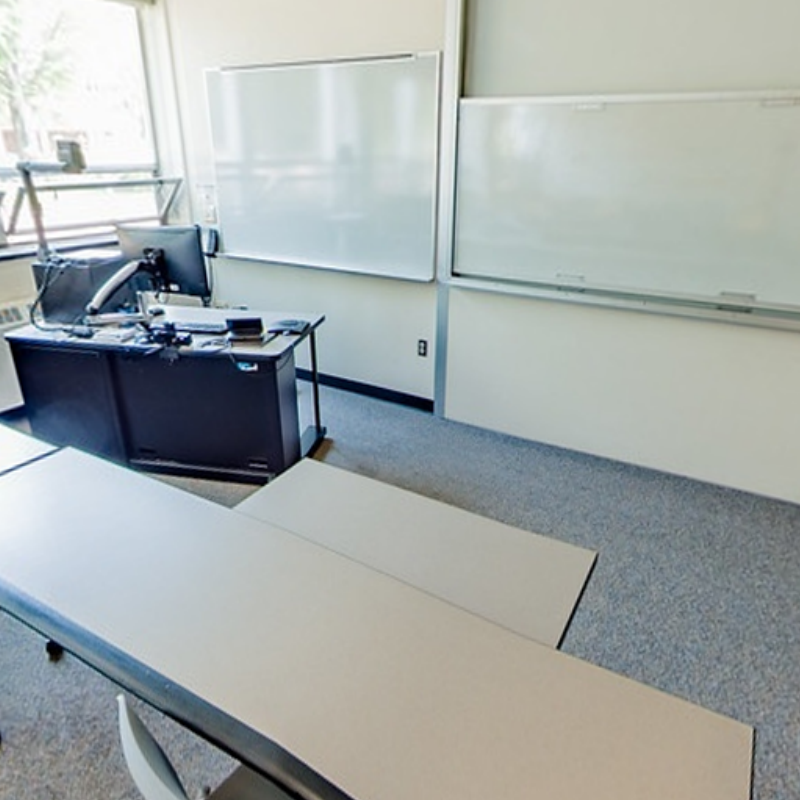} \hspace{-1mm} &
				\includegraphics[width=0.149\textwidth]{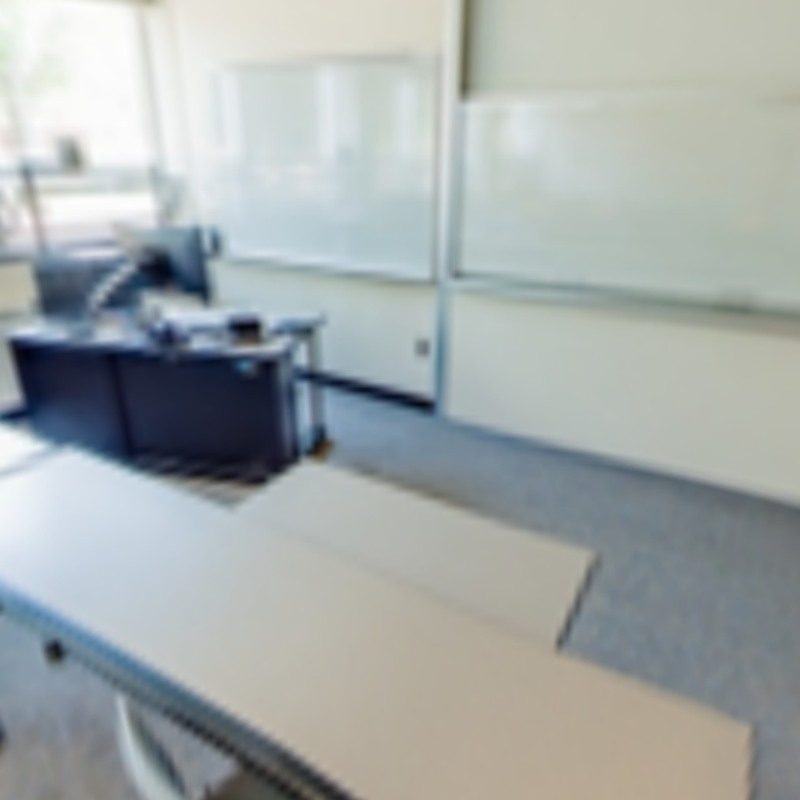} \hspace{-1mm} &
				\includegraphics[width=0.149\textwidth]{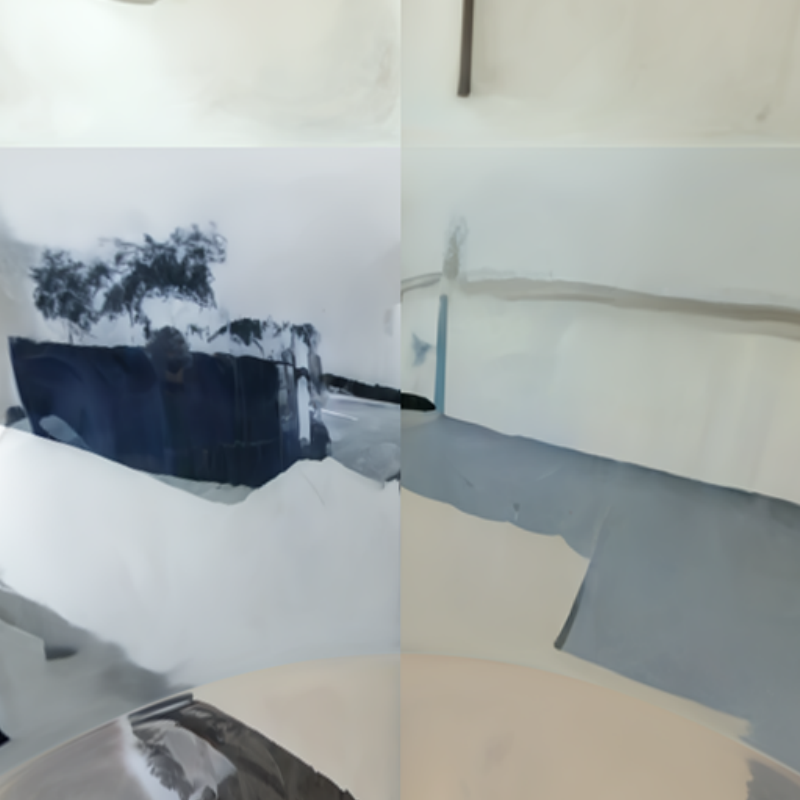} \hspace{-1mm} &
				\includegraphics[width=0.149\textwidth]{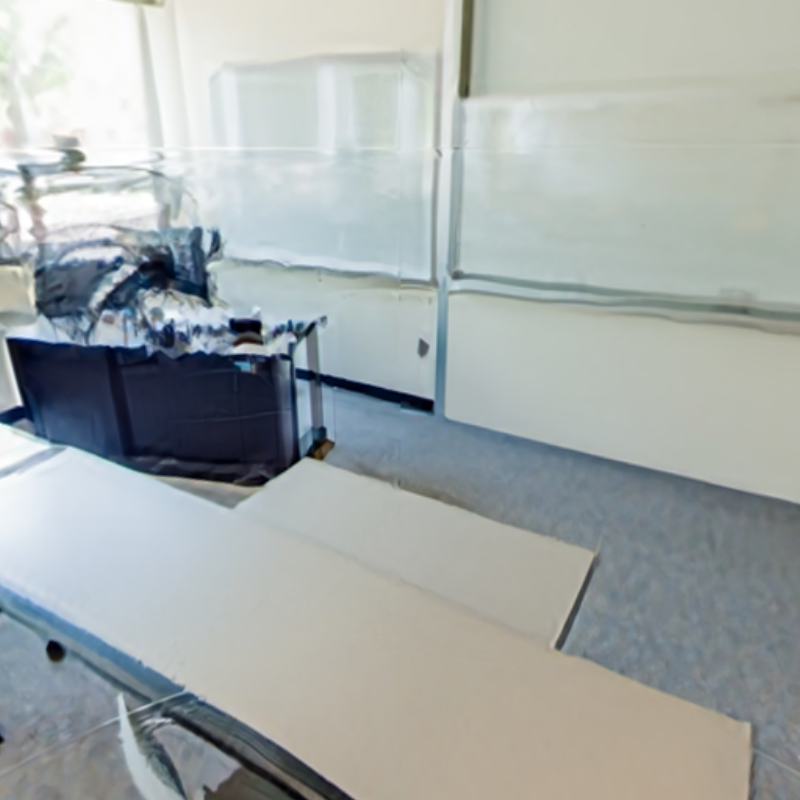} \hspace{-1mm} 
				\\
				HR \hspace{-1mm} &
				Bicubic \hspace{-1mm} &
				DPS \cite{chung2022diffusion} \hspace{-1mm} &
				DDRM \cite{kawar2022denoising} \hspace{-1mm}  
				\\
				PSNR/SSIM \hspace{-1mm} &
				24.65dB/0.7753 \hspace{-1mm} &
				22.57dB/0.7188 \hspace{-1mm} &
				25.36dB/0.8029 \hspace{-1mm} 
				\\
				\includegraphics[width=0.149\textwidth]{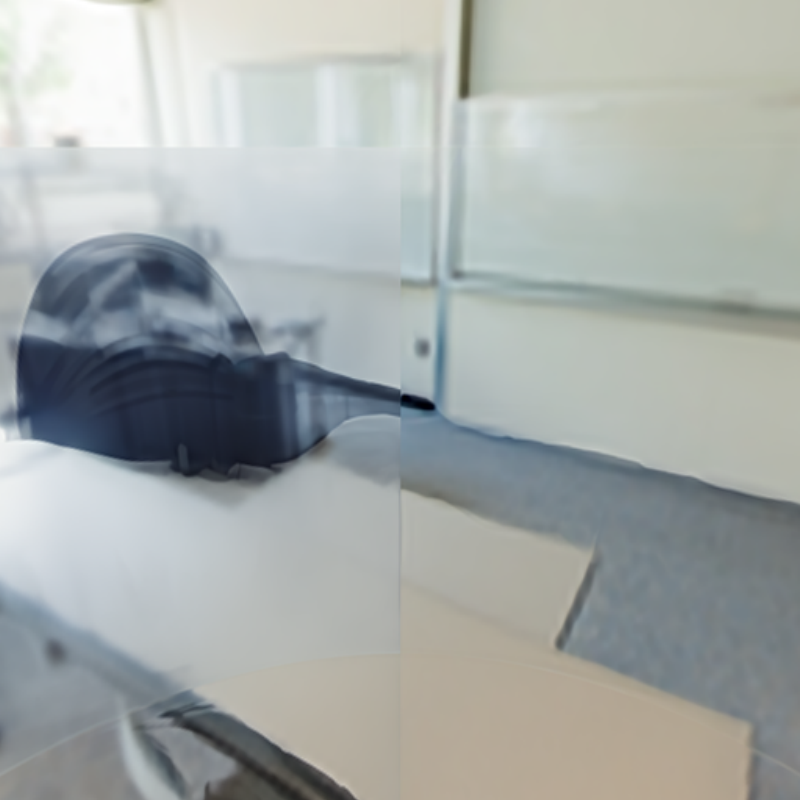} \hspace{-1mm} &
				\includegraphics[width=0.149\textwidth]{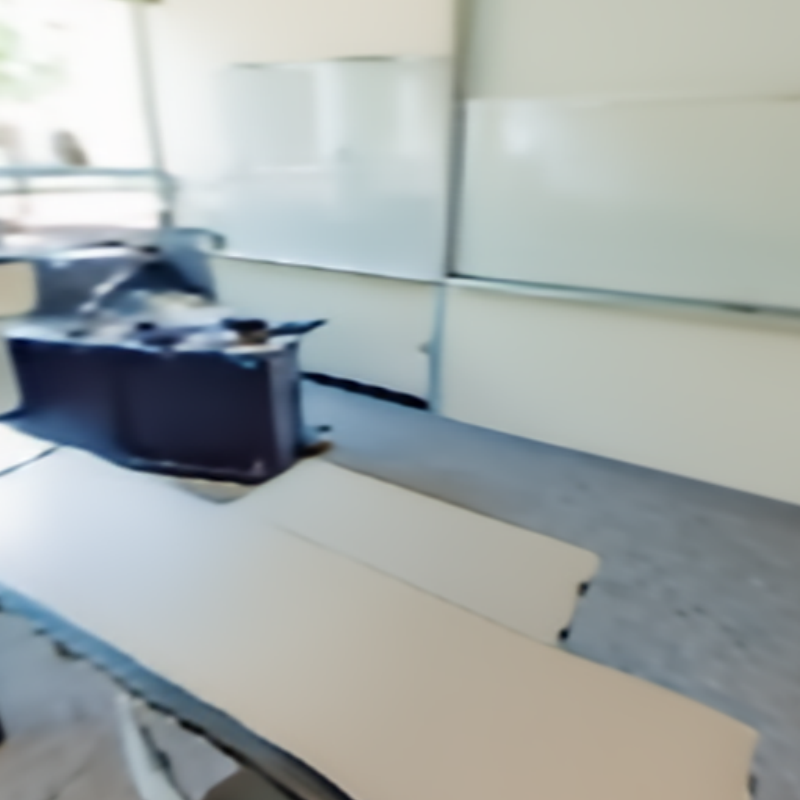} \hspace{-1mm} &
				\includegraphics[width=0.149\textwidth]{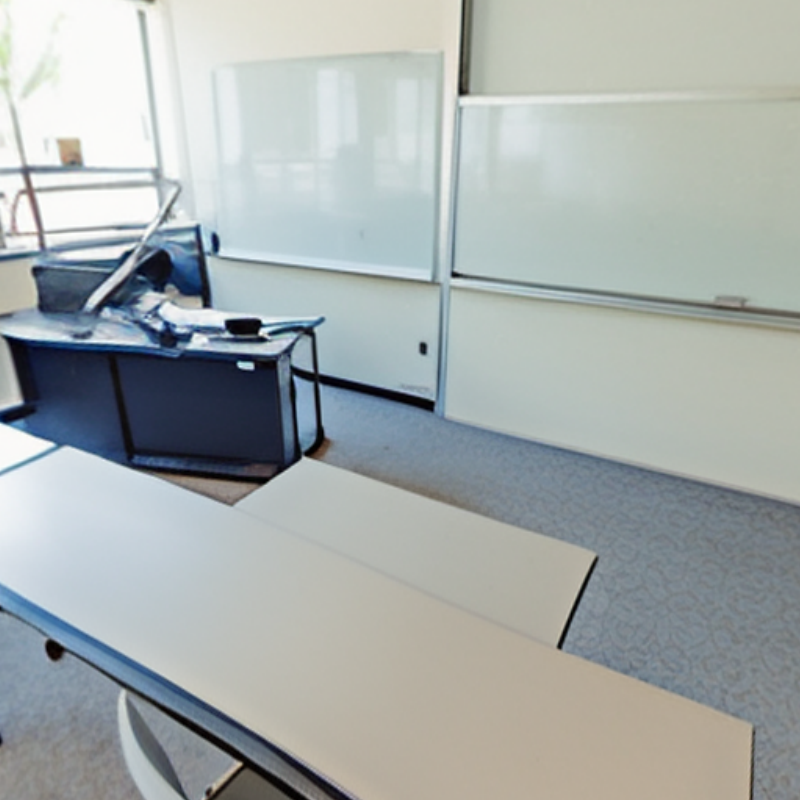} \hspace{-1mm} &
				\includegraphics[width=0.149\textwidth]{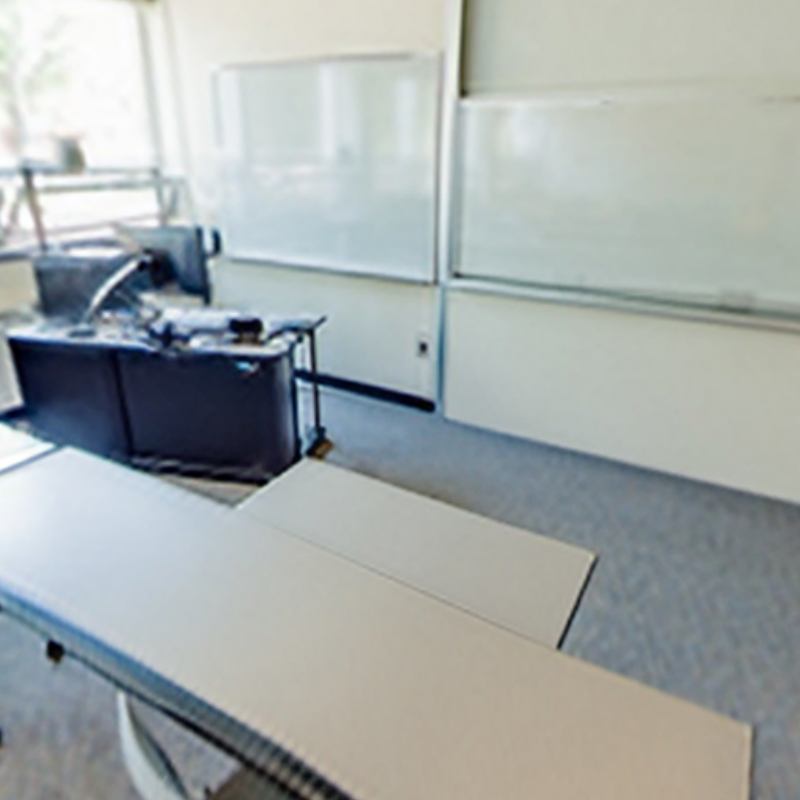} \hspace{-1mm}  
				\\ 
				GDP \cite{fei2023generative} \hspace{-1mm} &
				DiffIR \cite{Xia_Zhang_Wang_Wang_Wu_Tian_Yang_Gool} \hspace{-1mm} &
				StableSR \cite{StableSR_Wang_Yue_Zhou_Chan_Loy_2023} \hspace{-1mm} &
				\textbf{OmniSSR}(ours) \hspace{-1mm}
				\\
				23.02dB/0.7525 \hspace{-1mm} &
				23.48dB/0.7799 \hspace{-1mm} &
				24.62dB/0.7981 \hspace{-1mm} &
				26.53dB/0.8265 \hspace{-1mm} 
			\end{tabular}
		\end{adjustbox}
		\\ 
		
	\end{tabular}
	\caption{Visualized comparison of $\times$2 and $\times$4 SR results on SUN 360 testset. 001 and 009 is the id number in testset filenames. We also calculate the PSNR and SSIM to HR ground truth of each SR result and downsampled image.}
	\label{fig:visual}
	\vspace{-7mm}
\end{figure*}

\begin{figure*}[h]
	\scriptsize
	\centering
	\begin{tabular}{l}
		\hspace{-0.42cm}
		\begin{adjustbox}{valign=t}
			\begin{tabular}{c}
				\includegraphics[width=0.260\textwidth]{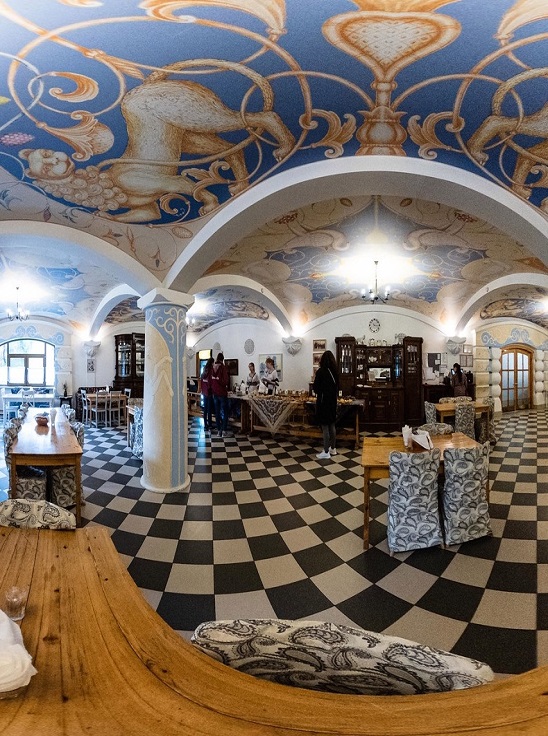}
				\vspace{1.5mm}
				\\
				ODI-SR ($\times$2): 067
			\end{tabular}
		\end{adjustbox}
		\hspace{-2mm}
		\begin{adjustbox}{valign=t}
			\begin{tabular}{cccc}
				\includegraphics[width=0.149\textwidth]{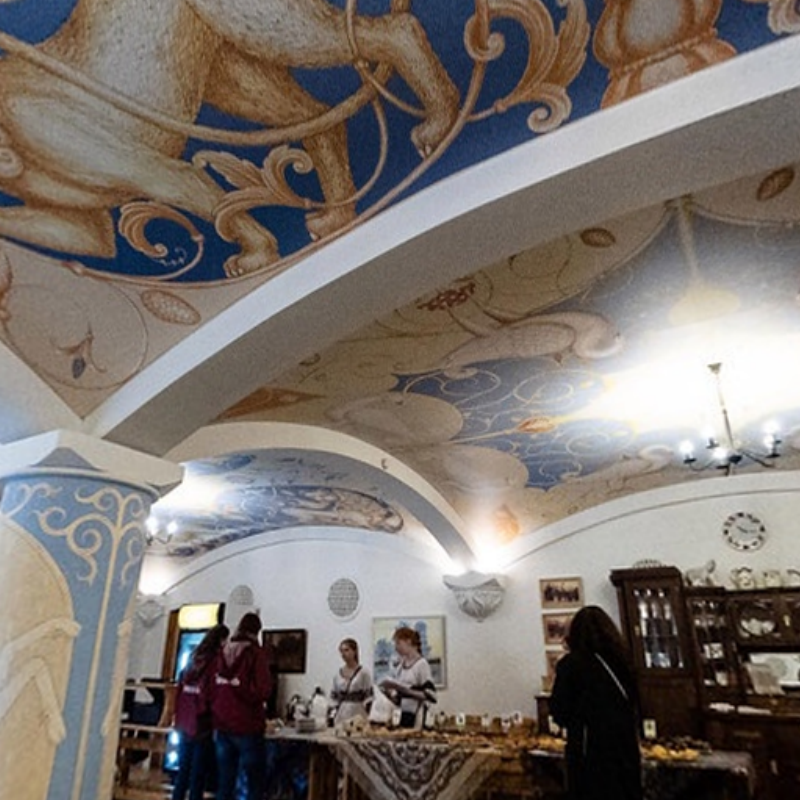} \hspace{-1mm} &
				\includegraphics[width=0.149\textwidth]{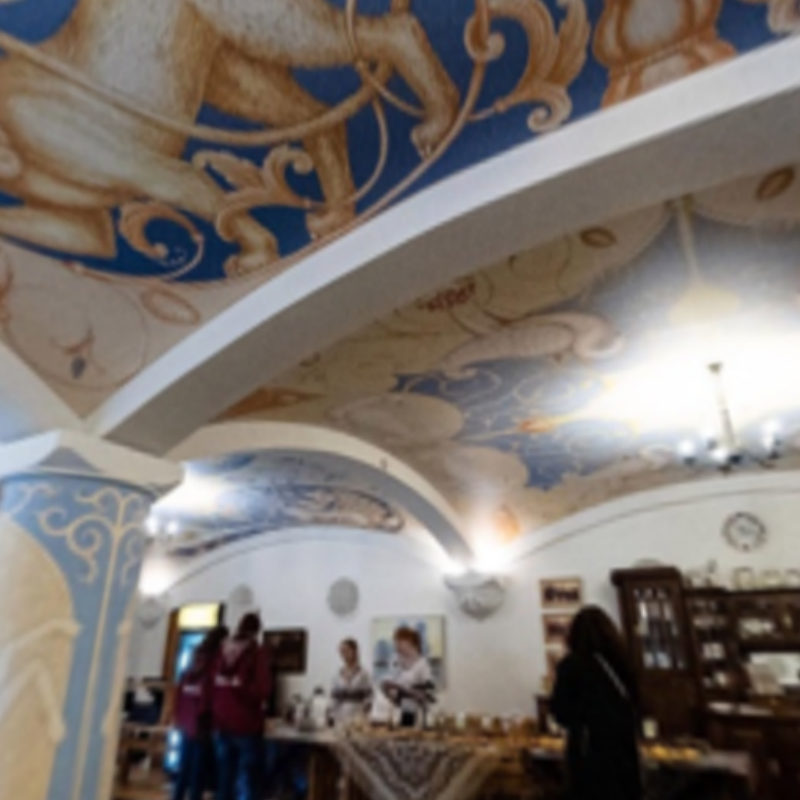} \hspace{-1mm} &
				\includegraphics[width=0.149\textwidth]{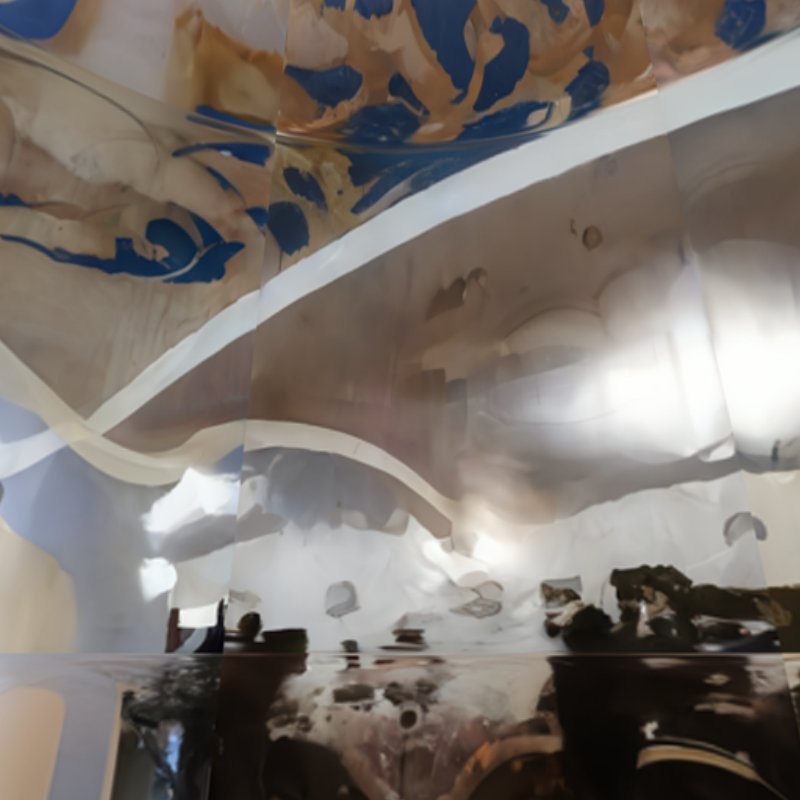} \hspace{-1mm} &
				\includegraphics[width=0.149\textwidth]{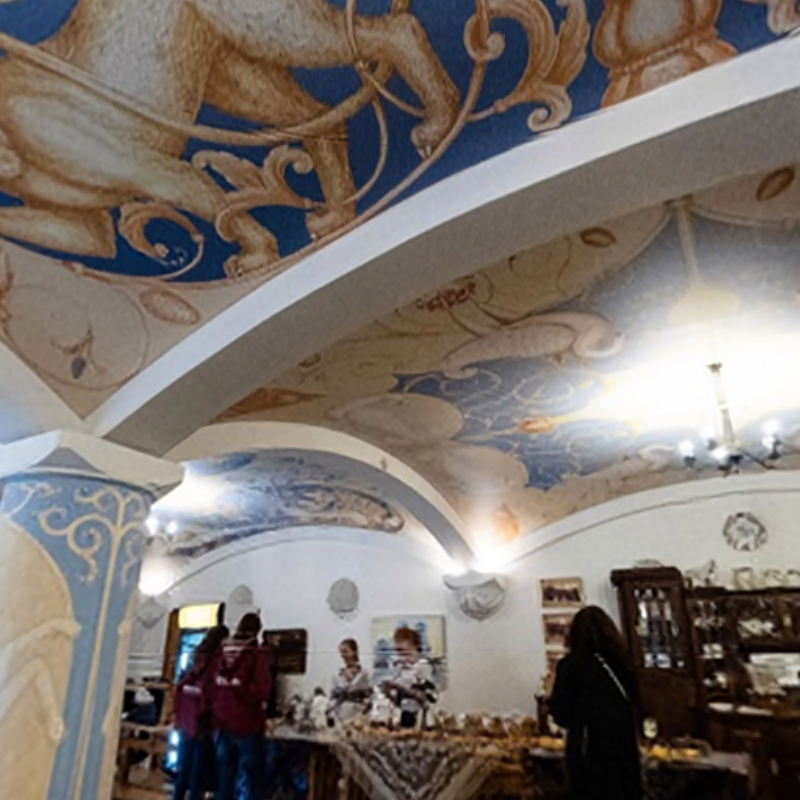} \hspace{-1mm} 
				\\
				HR \hspace{-1mm} &
				Bicubic \hspace{-1mm} &
				DPS \cite{chung2022diffusion} \hspace{-1mm} &
				DDRM \cite{kawar2022denoising} \hspace{-1mm} 
				\\ 
				PSNR/SSIM \hspace{-1mm} &
				27.67dB/0.8095 \hspace{-1mm} &
				22.93dB/0.5653 \hspace{-1mm} &
				29.91dB/0.8809 \hspace{-1mm} 
				\\
				\includegraphics[width=0.149\textwidth]{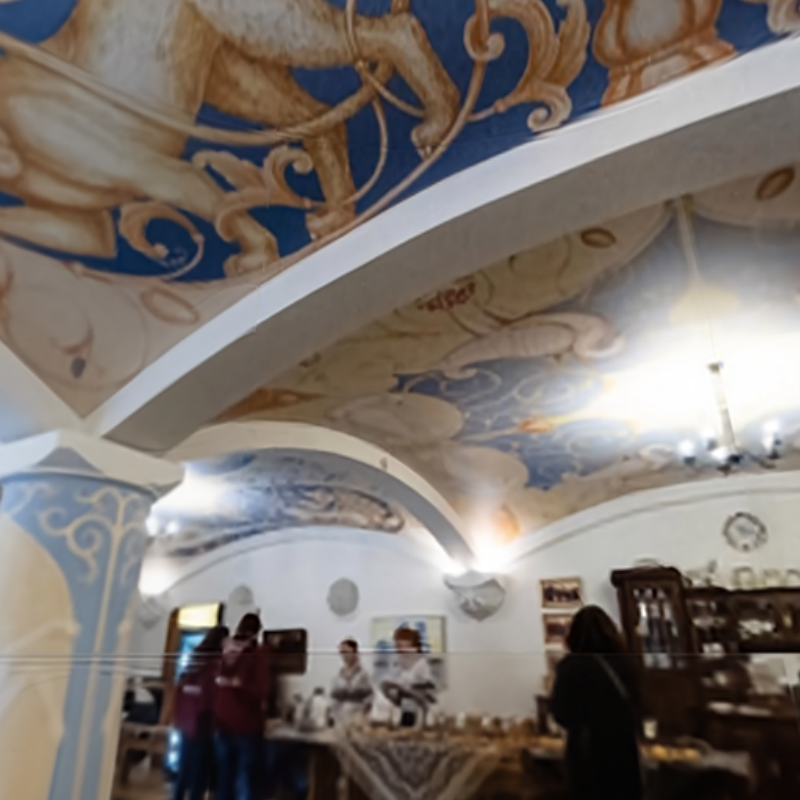} \hspace{-1mm} &
				\includegraphics[width=0.149\textwidth]{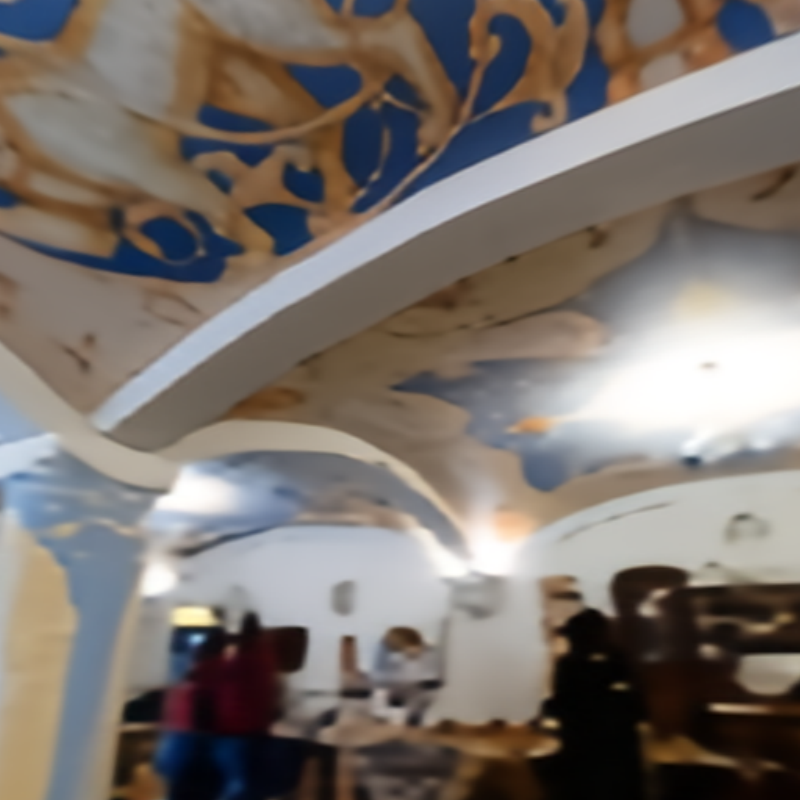} \hspace{-1mm} &
				\includegraphics[width=0.149\textwidth]{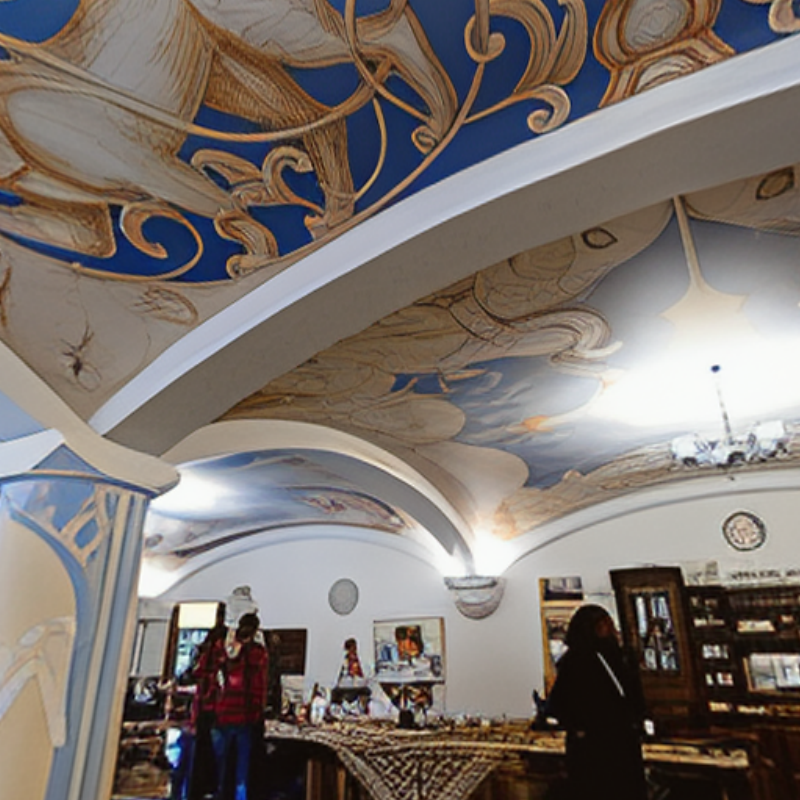} \hspace{-1mm} &
				\includegraphics[width=0.149\textwidth]{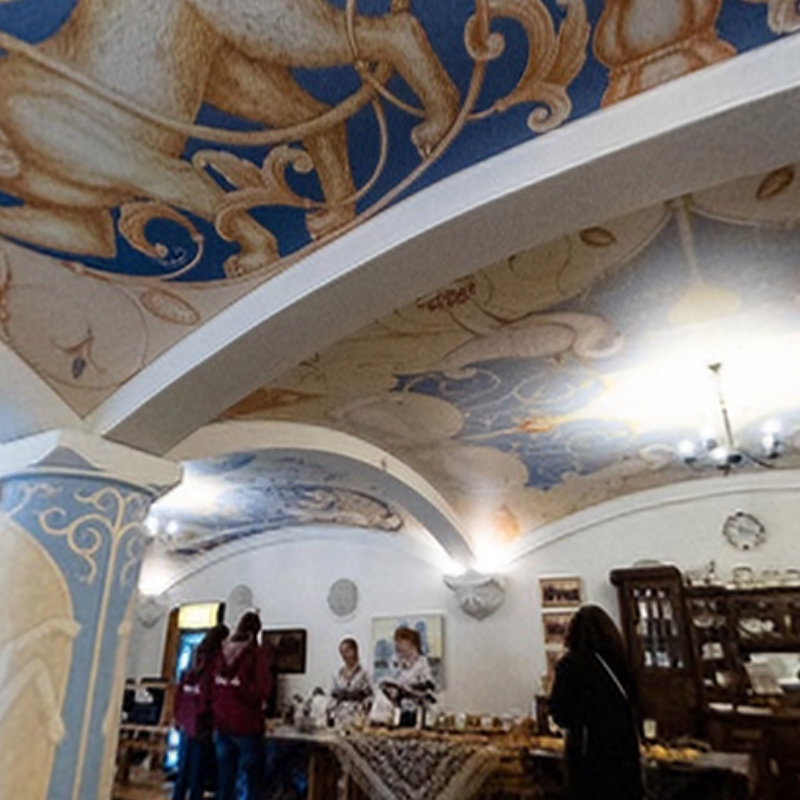} \hspace{-1mm}  
				\\ 
				GDP \cite{fei2023generative} \hspace{-1mm} &
				DiffIR \cite{Xia_Zhang_Wang_Wang_Wu_Tian_Yang_Gool} \hspace{-1mm} &
				StableSR \cite{StableSR_Wang_Yue_Zhou_Chan_Loy_2023} \hspace{-1mm} &
				\textbf{OmniSSR}(ours) \hspace{-1mm} 
				\\
				28.51dB/0.8258 \hspace{-1mm} &
				22.65dB/0.6248 \hspace{-1mm} &
				21.80dB/0.5892 \hspace{-1mm} &
				29.99dB/0.8798 \hspace{-1mm} 
			\end{tabular}
		\end{adjustbox}
		\vspace{2mm}
		
		\\ 
		\hspace{-0.42cm}
		\begin{adjustbox}{valign=t}
			\begin{tabular}{c}
				\includegraphics[width=0.260\textwidth]{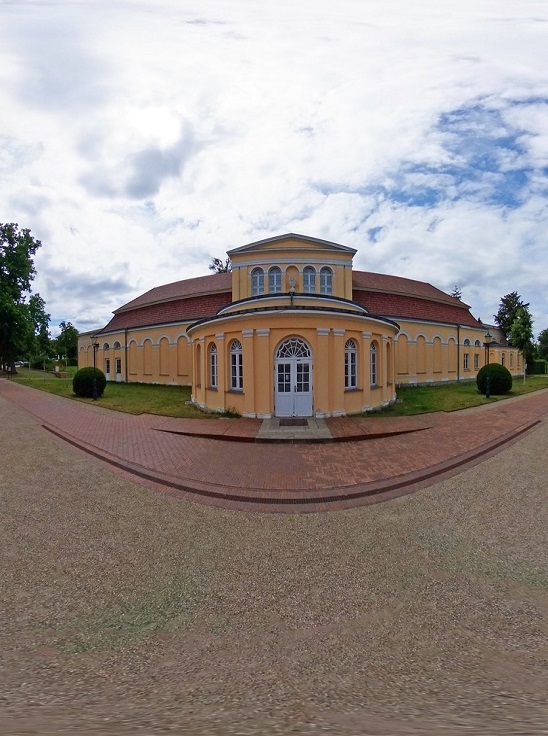}
				\vspace{1.5mm}
				\\
				ODI-SR ($\times$4): 049
			\end{tabular}
		\end{adjustbox}
		\hspace{-2mm}
		\begin{adjustbox}{valign=t}
			\begin{tabular}{cccc}
				\includegraphics[width=0.149\textwidth]{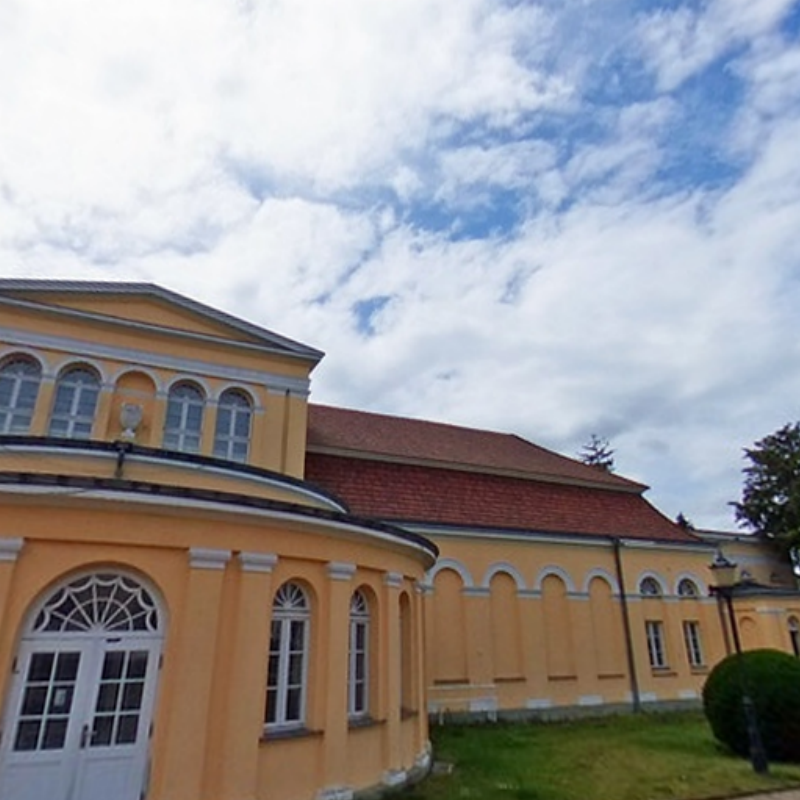} \hspace{-1mm} &
				\includegraphics[width=0.149\textwidth]{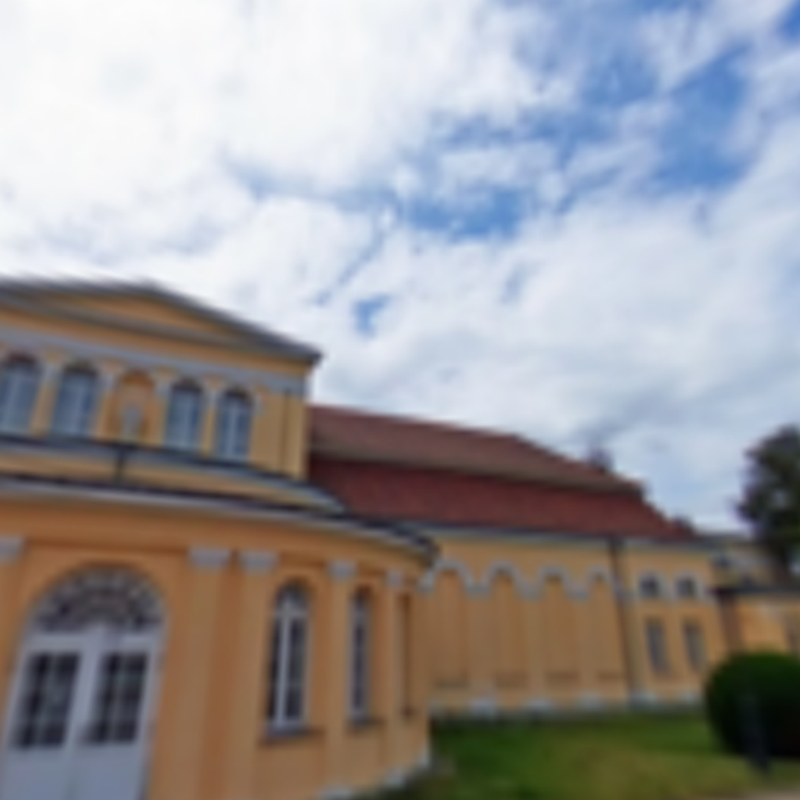} \hspace{-1mm} &
				\includegraphics[width=0.149\textwidth]{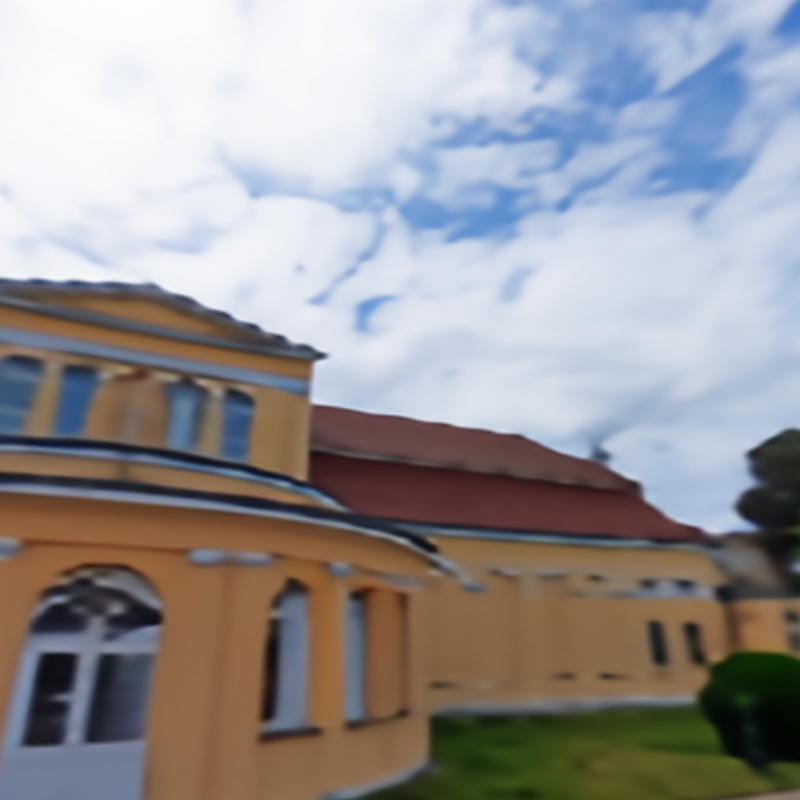} \hspace{-1mm} &
				\includegraphics[width=0.149\textwidth]{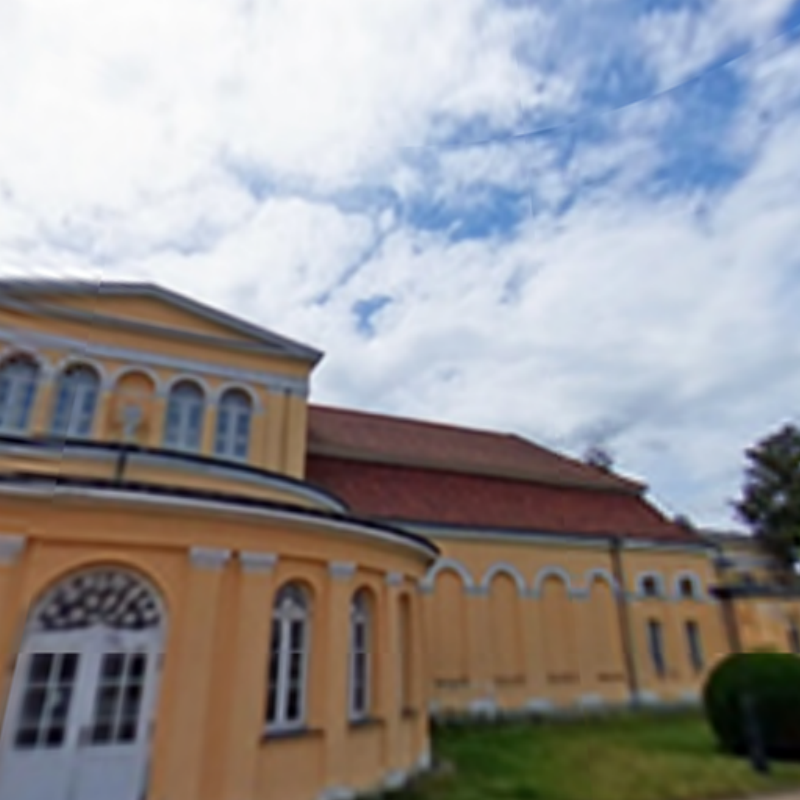} \hspace{-1mm} 
				\\
				HR \hspace{-1mm} &
				Bicubic \hspace{-1mm} &
				DPS \cite{chung2022diffusion} \hspace{-1mm} &
				DDRM \cite{kawar2022denoising} \hspace{-1mm}  
				\\
				PSNR/SSIM \hspace{-1mm} &
				25.44dB/0.7536 \hspace{-1mm} &
				26.21dB/0.7574 \hspace{-1mm} &
				27.12dB/0.8129 \hspace{-1mm} 
				\\
				\includegraphics[width=0.149\textwidth]{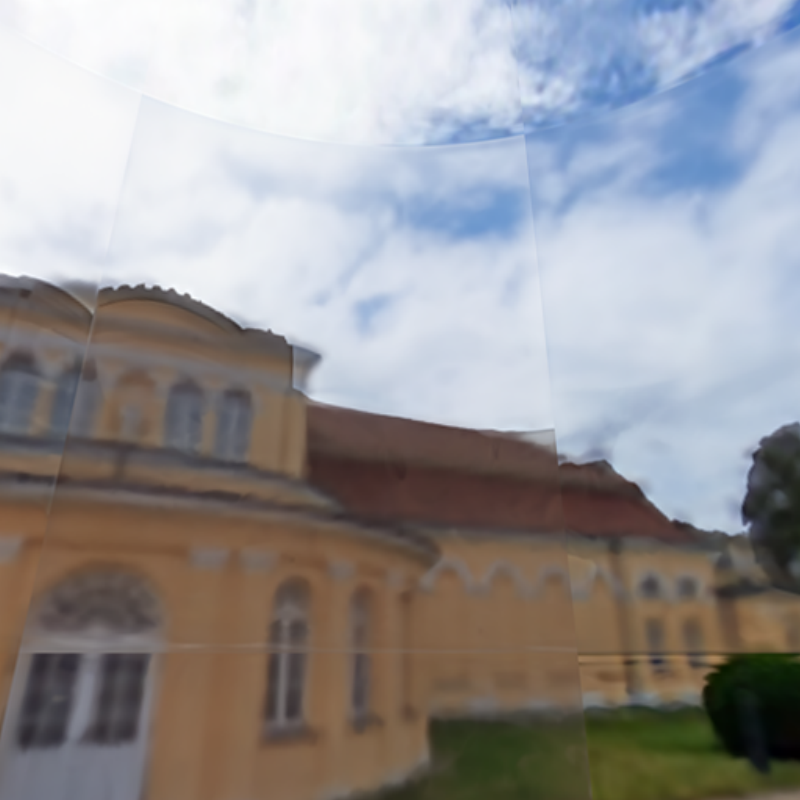} \hspace{-1mm} &
				\includegraphics[width=0.149\textwidth]{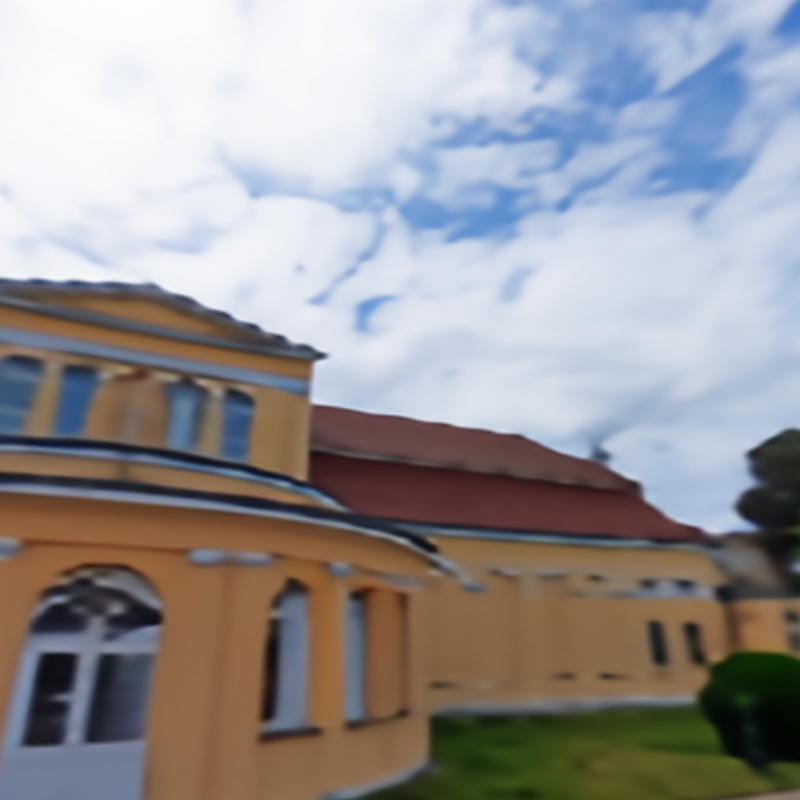} \hspace{-1mm} &
				\includegraphics[width=0.149\textwidth]{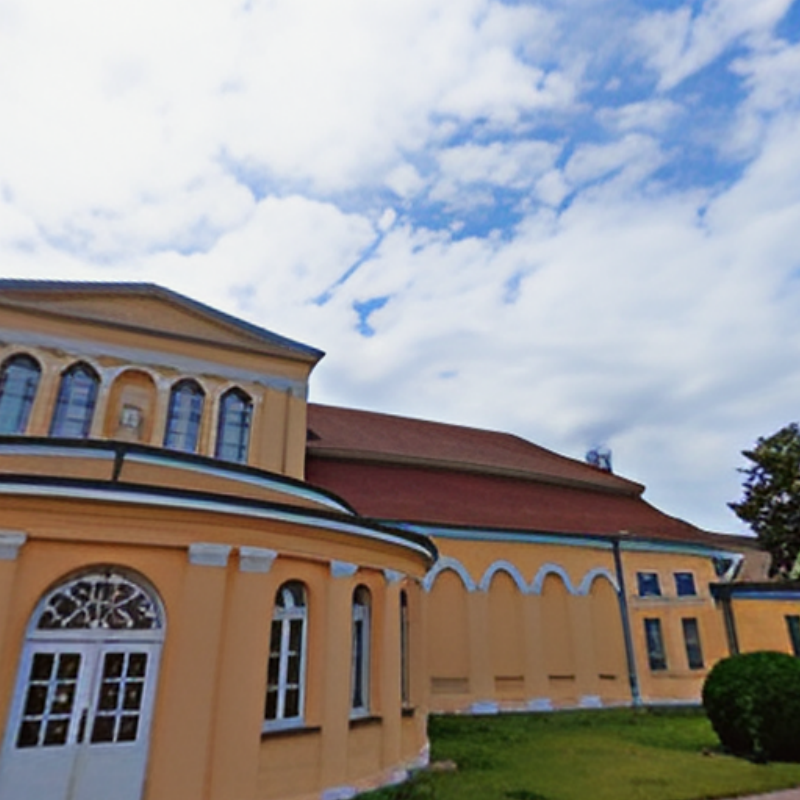} \hspace{-1mm} &
				\includegraphics[width=0.149\textwidth]{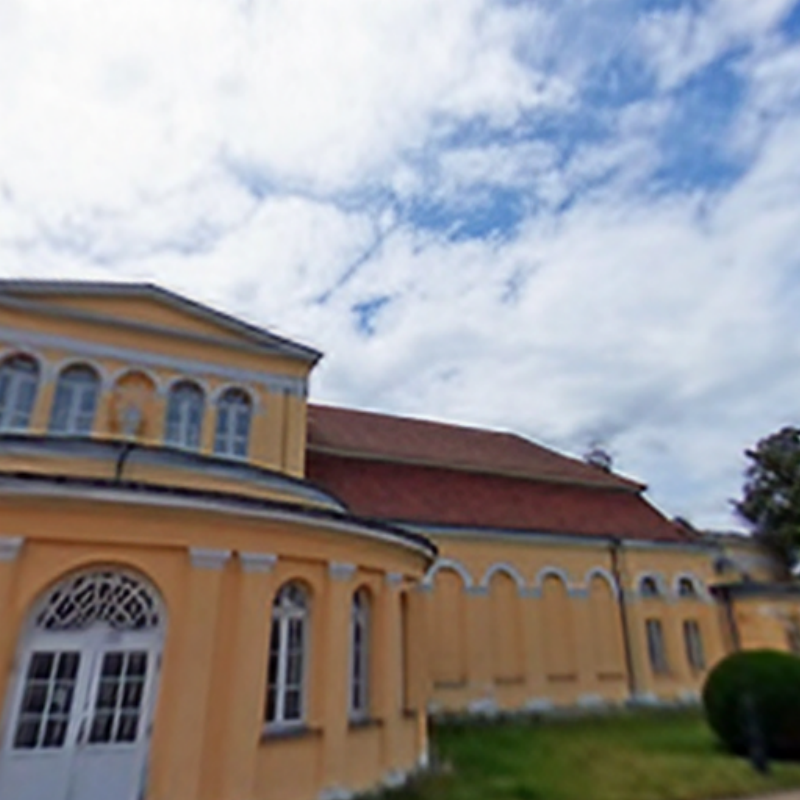} \hspace{-1mm}  
				\\ 
				GDP \cite{fei2023generative} \hspace{-1mm} &
				DiffIR \cite{Xia_Zhang_Wang_Wang_Wu_Tian_Yang_Gool} \hspace{-1mm} &
				StableSR \cite{StableSR_Wang_Yue_Zhou_Chan_Loy_2023} \hspace{-1mm} &
				\textbf{OmniSSR}(ours) \hspace{-1mm}
				\\
				24.15dB/0.7179 \hspace{-1mm} &
				24.25dB/0.7594 \hspace{-1mm} &
				21.80dB/0.5892 \hspace{-1mm} &
				27.20dB/0.8168 \hspace{-1mm} 
			\end{tabular}
		\end{adjustbox}
		\\ 
		
	\end{tabular}
	\caption{Visualized comparison of $\times$2 and $\times$4 SR results on ODI-SR test set. 067 and 049 are the id numbers in test set filenames. We also calculate the PSNR and SSIM between ground truth and each SR result as well as downsampled image.}
	\label{fig:visual_odisr}
	\vspace{-5mm}
\end{figure*}

\textbf{Qualitative results}
are shown in Fig.~\ref{fig:visual} and Fig.~\ref{fig:visual_odisr}, which illustrates the visualization of SR results on SUN test set and ODI-SR test set with $\times$2 and $\times$4 scales, by different methods.
The visual results indicate that our OmniSSR exhibits superior capability for detail recovery compared to other methods, particularly evident in textual elements (e.g., the text "flapping" in upper part of Fig.~\ref{fig:visual}), complex objects (e.g., the black desk with a screen in lower part of Fig.~\ref{fig:visual}, patterns above the white door in lower part of Fig.~\ref{fig:visual_odisr}), and small-scale objects (e.g., the person and clock behind the desk in upper part of Fig.~\ref{fig:visual_odisr}). OmniSSR demonstrates the ability to recover highly detailed and realistic visual effects from TP images.

\subsection{Comparison with end-to-end supervised methods}
The experiments of comparison in Sec.~\ref{sec:eval} are mainly focused on zero-shot image super-resolution methods, and supervised single image super-resolution methods, where the approaches are not trained or fine-tuned on omnidirectional images. In this part, we will compare OmniSSR to supervised end-to-end methods with end-to-end training on ODI datasets, including SwinIR and OSRT. Besides the main metrics in Sec.~\ref{sec:eval}, we also use NIQE~\cite{niqe} and DISTS~\cite{dists_ding2020image} to evaluate the visual perception of SR outputs. Results are presented in Tab.~\ref{tab:supervised}, which shows that although our OmniSSR exhibits inferior fidelity metrics compared to end-to-end supervised methods trained directly on ODI datasets, it demonstrates notable improvements in the visual quality and authenticity of super-resolved images. Notably, end-to-end methods often produce smoothed reconstructions with distortions, whereas our approach preserves finer details and adheres more closely to the realistic distribution.
Considering that our method has never been trained or tuned on ODI datasets, nor having omnidirectional images prior, this result is acceptable.
\begin{table}[h] 
	\centering
	\scriptsize
     \caption{Comparison on $\times$4 SR task with supervised methods trained on ODI-SR dataset, including SwinIR and OSRT. The best results are shown in \textbf{Bold}.}
         \vspace{-8pt}

		\begin{tabular}{c|c|cccccc}
			\toprule
			Method & Dataset & WS-PSNR$\uparrow$ & WS-SSIM$\uparrow$ & FID$\downarrow$ & LPIPS$\downarrow$ & NIQE$\downarrow$ & DISTS$\downarrow$  \\ \midrule
			SwinIR~\cite{liang2021swinir}                             & \multirow{3}{*}{ODI-SR} & 26.76   & 0.7620    & 27.94   & 0.3321 & 5.3961 & 0.1710 \\ 
			OSRT~\cite{osrt_Yu_Wang_Cao_Li_Shan_Dong_2023}                     &                       & \textbf{26.89}   & \textbf{0.7646}    & \textbf{27.39}   & 0.3258 & 5.4364 & 0.1695 \\
			OmniSSR             &              & 25.77   & 0.7279    & 30.97   & \textbf{0.2977} & \textbf{5.2891}  & \textbf{0.1541}\\ \hline
			SwinIR~\cite{liang2021swinir}                              & \multirow{3}{*}{SUN 360}               & 26.02   & 0.7692    & 39.90   & 0.3419 & 5.2440 & 0.1325 \\
            OSRT~\cite{osrt_Yu_Wang_Cao_Li_Shan_Dong_2023}                     &               & \textbf{26.33}   & \textbf{0.7766}    & 39.22   & 0.3364 & 5.2984 & 0.1312  \\ 
			OmniSSR                    &            & 26.01   & 0.7481    & \textbf{34.58}   & \textbf{0.2963} & \textbf{5.1329} & \textbf{0.1299} \\
            \bottomrule
		\end{tabular}
	\label{tab:supervised}
	\vspace{-3mm}
\end{table}


\subsection{Ablation Studies}
\label{sebsec: ablation study}
We first sequentially validate the performance improvement of the proposed strategy in OmniSSR including input image type, 
OTII and GD correction, on the ODI-SR test-set with $\times$2 SR task, thereby demonstrating the significance of these strategies.
The details are demonstrated as follows:

1) we do not use any proposed strategy in the SR task, which is equivalent to the vanilla StableSR baseline;

2) we transform the degraded ERP image to TP images and feed them separately into StableSR pipeline, instead of directly inputting ERP images; 

3) based on 2), we add OTII strategy during the denoising process of SD (Algo.~\ref{alg:sdsr+rnd} line~\ref{line:otii});

4) based on 2), we add GD correction at the \textit{post-processing} stage (Algo.~\ref{alg:pipeline} line~\ref{line:gd-post}) of the overall pipeline;

5) based on 3) and 4), we add GD correction at \textit{every step} and \textit{post-processing} stage of sampling, to improve the consistency of the restored result. 

Note that the execution of GD correction requires the execution of OTII in the denoising process simultaneously, there is no scenario where only GD correction is executed without the execution of OTII in the denoising process.
\begin{table}[h]
\centering
\scriptsize
\vspace{-3mm}
\caption{Ablation studies of OmniSSR on input type, OTII, and GD correction, on the test set of the ODI-SR dataset. Best results are shown in \textbf{Bold}.}
\label{tab:ablation}
\begin{tabular}{c|c|c|cccc}
\toprule
Input type & OTII & GD Correction & WS-PSNR$\uparrow$ & WS-SSIM$\uparrow$ & FID$\downarrow$&LPIPS$\downarrow$ \\
\hline
ERP&$\times$&$\times$&  22.69  & 0.6458 & 44.87&0.3039\\
\hline
TP&$\times$&$\times$ 
& 23.53 & 0.6849 & 43.91 & 0.3113 \\
\hline
TP&$\checkmark$ & $\times$
& 23.74 & 0.6847 & 65.35 & 0.3748 \\
\hline
TP&$\times$&$\checkmark$ 
(in post-process only)
&26.77 &0.8192&15.41&0.1691 \\
\hline
TP&$\checkmark$ & $\checkmark$ 
&\textbf{28.58}&\textbf{0.8540}&\textbf{13.01}&\textbf{0.1575}\\
\bottomrule
\end{tabular}
\vspace{-40pt}
\end{table}






\begin{table}[h]
\centering

\scriptsize
\caption{Results of pre-upsampling strategy on different scales, where ($x$,$y$) denotes bicubic-based upsampling at $x \times$ scale to ERP before ERP$\rightarrow$TP, and $y \times$ scale to TP before TP$\rightarrow$ERP transformation. Best results are shown in \textbf{Bold}.}
\label{tab:ablation_pu}
\begin{tabular}{c|c|c|c|c|c|c}
\toprule
ERP$\rightarrow$TP$\rightarrow$ERP & (1, 1) & (1, 4) & (4, 1) & (4, 2) & (2, 4) & (4, 4) \\ \hline
WS-PSNR$\uparrow$                    & 28.98  & 38.11  & 28.99  & 33.91  & 38.05  & \textbf{38.18}  \\ 
WS-SSIM$\uparrow$                    & 0.8859 & 0.9838 & 0.8862 & 0.9626 & 0.9837 & \textbf{0.9841} \\ \bottomrule
\end{tabular}
\vspace{-5mm}
\end{table}
Quantitative results of ablation studies are shown in Tab.~\ref{tab:ablation}. From the result shown below, we could come to the claim that the OTII helps improve the performance on the domain level, and the transformation between ERP and TP images provides information fusion among adjacent TP images. Our proposal of Gradient Decomposition corrects such restoration result, improving fidelity and realness significantly at the same time, and it would be better if it is applied at each step of the overall denoising pipeline. Tab.~\ref{tab:ablation_pu} shows the effect of mitigating information loss via proposed pre-upsampling strategy.

\begin{figure}
    \centering
    \includegraphics[width=0.9\linewidth]{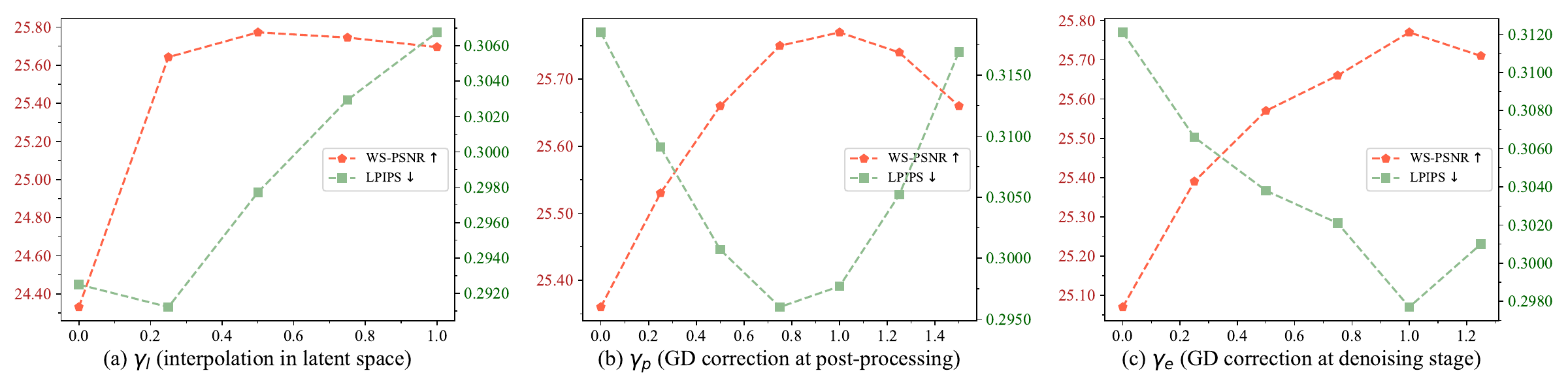}
    \vspace{-10pt}
    \caption{Ablation of choices on $\gamma_p$, $\gamma_e$ and $\gamma_l$. For better readability, WS-PSNR and LPIPS are chosen as evaluation metrics for fidelity and visual quality, respectively, to demonstrate the performance under different choices of the gamma parameter. We illustrate the results of (a) $\gamma_p$ and $\gamma_e$ fixed, while adjusting $\gamma_l$; (b) $\gamma_e$ and $\gamma_l$ fixed, while adjusting $\gamma_p$; (c) $\gamma_p$ and $\gamma_l$ fixed, while adjusting $\gamma_e$. It can be observed that when $\gamma_p=1$, $\gamma_e=1$, and $\gamma_l=0.5$, OmniSSR achieves the relatively best performance.}
    \label{fig:gamma}
    \vspace{-2mm}
\end{figure}
For $\gamma$ in the GD correction technique, we use grid search to obtain better results on ODI-SR dataset and $\times$4 SR task. Fig.~\ref{fig:gamma} shows performance on different choices of $\gamma_p$ in Algo.~\ref{alg:pipeline}
line~\ref{line:gd-post}, $\gamma_e$ in Algo.~\ref{alg:sdsr+rnd}
line~\ref{line:gammae}, and $\gamma_l$ in Algo.~\ref{alg:sdsr+rnd}
line~\ref{line:rnd-latent}. The entire ablation of $\gamma_p$, $\gamma_e$ and $\gamma_l$, with WS-PSNR, WS-SSIM, FID and LPIPS score all calculated and compared, will be provided in Supplementary Materials.

To evaluate the generalizability of our proposed modules, including Pre-Upsampling, OTII, and GD correction, we further conducted ablation studies on two super-resolution backbones, StableSR and SwinIR. The results underscore substantial performance enhancements facilitated by our modules across both backbones, which is provided in Supplementary Materials.

\section{Limitation and Discussion}
\label{sec:discussion}
Although OmniSSR bridges the gap between omnidirectional and planar images, achieving competitive performance and better visual results in ODISR, it still exhibits the following limitations: (1) The inference of the diffusion model requires a considerable amount of time, approximately 14 minutes per ERP-formatted omnidirectional image to be super-resolved into size $1024\times 2048$, making real-time super-resolution challenging; (2) Multiple conversions between ERP and TP are required in the pipeline, leading to improved performance but consuming additional inference time; (3) Further exploration of the convex optimization properties of GD correction is warranted, such as designing gradient term coefficients adaptive to reconstruction results and degradation types.

This study explores the application of image generation models to ODISR tasks. In future work, the framework behind OmniSSR can be extended beyond the confines of image super-resolution in a single scenario and venture into more complex ODI-based real-world scenarios. These include ODI editing, ODI inpainting, enhancing the quality of 3D Gaussian Splatting scenes~\cite{kerbl20233d,schonbein2014omnidirectional} obtained after super-resolving ERP images, as well as enhancing the quality of omnidirectional videos~\cite{wang2024360dvd}.

\section{Conclusion}
This paper leverages the image prior of Stable Diffusion (SD) and employs the Octadecaplex Tangent Information Interaction (OTII) to achieve \textit{zero-shot} omnidirectional image super-resolution. Additionally, we propose the Gradient Decomposition (GD) correction based on convex optimization algorithms to refine the initial super-resolution results, enhancing the fidelity and realness of the restored images. The superior performance of our proposed method, OmniSSR, is demonstrated on benchmark datasets. By bridging the gap between omnidirectional and planar images, we establish a training-free approach, mitigating the data demand and over-fitting associated with end-to-end training. The application scope of our method can be further extended to various applications, presenting potential value across multiple visual tasks.


%
%
\bibliographystyle{splncs04}
\bibliography{main}

\begin{thebibliography}{10}
\providecommand{\url}[1]{\texttt{#1}}
\providecommand{\urlprefix}{URL }
\providecommand{\doi}[1]{https://doi.org/#1}

\bibitem{10222760}
An, H., Zhang, X.: Perception-oriented omnidirectional image super-resolution based on transformer network. In: Proceedings of the IEEE International Conference on Image Processing (ICIP) (2023)

\bibitem{arican2011joint}
Arican, Z., Frossard, P.: Joint registration and super-resolution with omnidirectional images. IEEE Transactions on Image Processing (TIP)  (2011)

\bibitem{Cao_2023_CVPR}
Cao, M., Mou, C., Yu, F., Wang, X., Zheng, Y., Zhang, J., Dong, C., Li, G., Shan, Y., Timofte, R., et~al.: Ntire 2023 challenge on 360deg omnidirectional image and video super-resolution: Datasets, methods and results. In: Proceedings of the IEEE/CVF Conference on Computer Vision and Pattern Recognition Workshops (CVPRW) (2023)

\bibitem{chan2022glean}
Chan, K.C., Xu, X., Wang, X., Gu, J., Loy, C.C.: Glean: Generative latent bank for image super-resolution and beyond. IEEE Transactions on Pattern Analysis and Machine Intelligence (TPAMI)  (2022)

\bibitem{Chen_Liu_Wang_2021}
Chen, Y., Liu, S., Wang, X.: Learning continuous image representation with local implicit image function. In: Proceedings of the IEEE/CVF Conference on Computer Vision and Pattern Recognition (CVPR) (2021)

\bibitem{Chen_2023_ICCV}
Chen, Z., Zhang, Y., Gu, J., Kong, L., Yang, X., Yu, F.: Dual aggregation transformer for image super-resolution. In: Proceedings of the IEEE/CVF international conference on computer vision (ICCV) (2023)

\bibitem{cheng2023hybrid}
Cheng, M., Ma, H., Ma, Q., Sun, X., Li, W., Zhang, Z., Sheng, X., Zhao, S., Li, J., Zhang, L.: Hybrid transformer and cnn attention network for stereo image super-resolution. In: Proceedings of the IEEE/CVF Conference on Computer Vision and Pattern Recognition (CVPR) (2023)

\bibitem{mou2022mlim}
Chong, M., Yanze, W., Xintao, W., Chao, D., Jian, Z., Ying, S.: Metric learning based interactive modulation for real-world super-resolution. In: Proceedings of the European Conference on Computer Vision (ECCV) (2022)

\bibitem{chung2022diffusion}
Chung, H., Kim, J., Mccann, M.T., Klasky, M.L., Ye, J.C.: Diffusion posterior sampling for general noisy inverse problems. arXiv preprint arXiv:2209.14687  (2022)

\bibitem{chung2022improving}
Chung, H., Sim, B., Ye, J.C.: Improving diffusion models for inverse problems using manifold constraints. In: Proceedings of the Advances in Neural Information Processing Systems (NeurIPS) (2022)

\bibitem{Chung_Ye_Milanfar_Delbracio}
Chung, H., Ye, J., Milanfar, P., Delbracio, M.: Prompt-tuning latent diffusion models for inverse problems. arXiv preprint arXiv:2310.01110  (2023)

\bibitem{coxeter1961introduction}
Coxeter, H.S.M.: Introduction to geometry. John Wiley \& Sons, Inc. (1961)

\bibitem{daras2021intermediate}
Daras, G., Dean, J., Jalal, A., Dimakis, A.: Intermediate layer optimization for inverse problems using deep generative models. In: Proceedings of the International Conference on Machine Learning (ICML) (2021)

\bibitem{Deng_Wang_Xu_Guo_Song_Yang_2021}
Deng, X., Wang, H., Xu, M., Guo, Y., Song, Y., Yang, L.: Lau-net: Latitude adaptive upscaling network for omnidirectional image super-resolution. In: Proceedings of the IEEE/CVF Conference on Computer Vision and Pattern Recognition (CVPR) (2021)

\bibitem{deng2022omnidirectional}
Deng, X., Wang, H., Xu, M., Li, L., Wang, Z.: Omnidirectional image super-resolution via latitude adaptive network. IEEE Transactions on Multimedia (TMM)  (2022)

\bibitem{dists_ding2020image}
Ding, K., Ma, K., Wang, S., Simoncelli, E.P.: Image quality assessment: Unifying structure and texture similarity. IEEE Transactions on Pattern Analysis and Machine Intelligence (TPAMI)  (2020)

\bibitem{dong2015image}
Dong, C., Loy, C.C., He, K., Tang, X.: Image super-resolution using deep convolutional networks. IEEE Transactions on Pattern Analysis and Machine Intelligence (TPAMI)  (2015)

\bibitem{duan2018perceptual}
Duan, H., Zhai, G., Min, X., Zhu, Y., Fang, Y., Yang, X.: Perceptual quality assessment of omnidirectional images. In: Proceedings of the IEEE International Symposium on Circuits and Systems (ISCAS) (2018)

\bibitem{Fakour-Sevom_Guldogan_Kamarainen_2018}
Fakour-Sevom, V., Guldogan, E., K{\"a}m{\"a}r{\"a}inen, J.K.: 360 panorama super-resolution using deep convolutional networks. In: Proceedings of the Int. Conf. on Computer Vision Theory and Applications (VISAPP) (2018)

\bibitem{fei2023generative}
Fei, B., Lyu, Z., Pan, L., Zhang, J., Yang, W., Luo, T., Zhang, B., Dai, B.: Generative diffusion prior for unified image restoration and enhancement. In: Proceedings of the IEEE/CVF Conference on Computer Vision and Pattern Recognition (CVPR) (2023)

\bibitem{gan}
Goodfellow, I., Pouget-Abadie, J., Mirza, M., Xu, B., Warde-Farley, D., Ozair, S., Courville, A., Bengio, Y.: Generative adversarial nets. In: Proceedings of the Advances in Neural Information Processing Systems (NeurIPS) (2014)

\bibitem{guo2024cas}
Guo, L., Tao, T., Cai, X., Zhu, Z., Huang, J., Zhu, L., Gu, Z., Tang, H., Zhou, R., Han, S., et~al.: Cas-diffcom: Cascaded diffusion model for infant longitudinal super-resolution 3d medical image completion. arXiv preprint arXiv:2402.13776  (2024)

\bibitem{fid}
Heusel, M., Ramsauer, H., Unterthiner, T., Nessler, B., Hochreiter, S.: Gans trained by a two time-scale update rule converge to a local nash equilibrium. In: Proceedings of the Advances in Neural Information Processing Systems (NeurIPS) (2017)

\bibitem{ho2020denoising}
Ho, J., Jain, A., Abbeel, P.: Denoising diffusion probabilistic models. In: Proceedings of the Advances in Neural Information Processing Systems (NeurIPS) (2020)

\bibitem{jiang2022reference}
Jiang, Y., Chan, K.C., Wang, X., Loy, C.C., Liu, Z.: Reference-based image and video super-resolution via $c^2$-matching. IEEE Transactions on Pattern Analysis and Machine Intelligence (TPAMI)  (2022)

\bibitem{kawar2022denoising}
Kawar, B., Elad, M., Ermon, S., Song, J.: Denoising diffusion restoration models. In: Proceedings of the ICLR Workshop on Deep Generative Models for Highly Structured Data (ICLRW) (2022)

\bibitem{kerbl20233d}
Kerbl, B., Kopanas, G., Leimk{\"u}hler, T., Drettakis, G.: 3d gaussian splatting for real-time radiance field rendering. ACM Transactions on Graphics (TOG)  (2023)

\bibitem{kim2023regularization}
Kim, J., Park, G.Y., Chung, H., Ye, J.C.: Regularization by texts for latent diffusion inverse solvers. arXiv preprint arXiv:2311.15658  (2023)

\bibitem{li2022d3c2}
Li, W., Chen, B., Zhang, J.: D3c2-net: Dual-domain deep convolutional coding network for compressive sensing. arXiv preprint arXiv:2207.13560  (2022)

\bibitem{li2022omnifusion}
Li, Y., Guo, Y., Yan, Z., Huang, X., Duan, Y., Ren, L.: Omnifusion: 360 monocular depth estimation via geometry-aware fusion. In: Proceedings of the IEEE/CVF Conference on Computer Vision and Pattern Recognition (CVPR) (2022)

\bibitem{liang2021swinir}
Liang, J., Cao, J., Sun, G., Zhang, K., Van~Gool, L., Timofte, R.: Swinir: Image restoration using swin transformer. In: Proceedings of the IEEE/CVF International Conference on Computer Vision Workshops (ICCVW) (2021)

\bibitem{liu2023residual}
Liu, J., Wang, Q., Fan, H., Wang, Y., Tang, Y., Qu, L.: Residual denoising diffusion models. arXiv preprint arXiv:2308.13712  (2023)

\bibitem{Lu_2022_CVPR}
Lu, Z., Li, J., Liu, H., Huang, C., Zhang, L., Zeng, T.: Transformer for single image super-resolution. In: Proceedings of the IEEE/CVF Conference on Computer Vision and Pattern Recognition Workshops (CVPRW) (2022)

\bibitem{Lugmayr_2020_CVPR_Workshops}
Lugmayr, A., Danelljan, M., Timofte, R.: Ntire 2020 challenge on real-world image super-resolution: Methods and results. In: Proceedings of the IEEE/CVF Conference on Computer Vision and Pattern Recognition Workshops (CVPRW) (2020)

\bibitem{menon2020pulse}
Menon, S., Damian, A., Hu, S., Ravi, N., Rudin, C.: Pulse: Self-supervised photo upsampling via latent space exploration of generative models. In: Proceedings of the IEEE/CVF Conference on Computer Vision and Pattern Recognition (CVPR) (2020)

\bibitem{niqe}
Mittal, A., Soundararajan, R., Bovik, A.C.: Making a “completely blind” image quality analyzer. IEEE Signal Processing Letters (SPL)  (2013)

\bibitem{9506233}
Nishiyama, A., Ikehata, S., Aizawa, K.: 360° single image super resolution via distortion-aware network and distorted perspective images. In: Proceedings of the IEEE International Conference on Image Processing (ICIP) (2021)

\bibitem{ozcinar2019super}
Ozcinar, C., Rana, A., Smolic, A.: Super-resolution of omnidirectional images using adversarial learning. In: Proceedings of the IEEE International Workshop on Multimedia Signal Processing (MMSPW) (2019)

\bibitem{pan2021exploiting}
Pan, X., Zhan, X., Dai, B., Lin, D., Loy, C.C., Luo, P.: Exploiting deep generative prior for versatile image restoration and manipulation. IEEE Transactions on Pattern Analysis and Machine Intelligence (TPAMI)  (2021)

\bibitem{Rombach_Blattmann_Lorenz_Esser_Ommer_2022}
Rombach, R., Blattmann, A., Lorenz, D., Esser, P., Ommer, B.: High-resolution image synthesis with latent diffusion models. In: Proceedings of the IEEE/CVF Conference on Computer Vision and Pattern Recognition (CVPR) (2022)

\bibitem{Rout_Raoof_Daras_Caramanis_Dimakis_Shakkottai_2023}
Rout, L., Raoof, N., Daras, G., Caramanis, C., Dimakis, A., Shakkottai, S.: Solving linear inverse problems provably via posterior sampling with latent diffusion models. In: Proceedings of the Advances in Neural Information Processing Systems (NeurIPS) (2023)

\bibitem{sr3}
Saharia, C., Ho, J., Chan, W., Salimans, T., Fleet, D.J., Norouzi, M.: Image super-resolution via iterative refinement. IEEE Transactions on Pattern Analysis and Machine Intelligence (TPAMI)  (2022)

\bibitem{schonbein2014omnidirectional}
Sch{\"o}nbein, M., Geiger, A.: Omnidirectional 3d reconstruction in augmented manhattan worlds. In: Proceedings of the IEEE/RSJ International Conference on Intelligent Robots and Systems (IROS) (2014)

\bibitem{Song_Zhang_Yin_Mardani_Liu_Kautz_Chen_Vahdat}
Song, J., Zhang, Q., Yin, H., Mardani, M., Liu, M.Y., Kautz, J., Chen, Y., Vahdat, A.: Loss-guided diffusion models for plug-and-play controllable generation. In: Proceedings of the International Conference on Machine Learning (ICML) (2023)

\bibitem{song2020score}
Song, Y., Sohl-Dickstein, J., Kingma, D.P., Kumar, A., Ermon, S., Poole, B.: Score-based generative modeling through stochastic differential equations. In: Proceedings of the International Conference on Learning Representations (ICLR) (2020)

\bibitem{sun2023opdn}
Sun, X., Li, W., Zhang, Z., Ma, Q., Sheng, X., Cheng, M., Ma, H., Zhao, S., Zhang, J., Li, J., et~al.: Opdn: Omnidirectional position-aware deformable network for omnidirectional image super-resolution. In: Proceedings of the IEEE/CVF Conference on Computer Vision and Pattern Recognition Workshops (CVPRW) (2023)

\bibitem{WS_PSNR_Sun_Lu_Yu_2017}
Sun, Y., Lu, A., Yu, L.: Weighted-to-spherically-uniform quality evaluation for omnidirectional video. IEEE Signal Processing Letters (SPL)  (2017)

\bibitem{Vaswani_Shazeer_Parmar_Uszkoreit_Jones_Gomez_Kaiser_Polosukhin_2017}
Vaswani, A., Shazeer, N., Parmar, N., Uszkoreit, J., Jones, L., Gomez, A., Kaiser, L., Polosukhin, I.: Attention is all you need. In: Proceedings of the Advances in Neural Information Processing Systems (NeurIPS) (2017)

\bibitem{StableSR_Wang_Yue_Zhou_Chan_Loy_2023}
Wang, J., Yue, Z., Zhou, S., Chan, K., Loy, C.: Exploiting diffusion prior for real-world image super-resolution. arXiv preprint arXiv:2305.07015  (2023)

\bibitem{wang2024360dvd}
Wang, Q., Li, W., Mou, C., Cheng, X., Zhang, J.: 360dvd: Controllable panorama video generation with 360-degree video diffusion model. In: Proceedings of the IEEE/CVF Conference on Computer Vision and Pattern Recognition (CVPR) (2024)

\bibitem{wang2021real}
Wang, X., Xie, L., Dong, C., Shan, Y.: Real-esrgan: Training real-world blind super-resolution with pure synthetic data. In: Proceedings of the IEEE/CVF International Conference on Computer Vision (ICCV) (2021)

\bibitem{wang2018esrgan}
Wang, X., Yu, K., Wu, S., Gu, J., Liu, Y., Dong, C., Qiao, Y., Change~Loy, C.: Esrgan: Enhanced super-resolution generative adversarial networks. In: Proceedings of the European Conference on Computer Vision Workshops (ECCVW) (2018)

\bibitem{wang2022zero}
Wang, Y., Yu, J., Zhang, J.: Zero-shot image restoration using denoising diffusion null-space model. In: Proceedings of the International Conference on Learning Representations (ICLR) (2022)

\bibitem{Xia_Zhang_Wang_Wang_Wu_Tian_Yang_Gool}
Xia, B., Zhang, Y., Wang, S., Wang, Y., Wu, X., Tian, Y., Yang, W., Van~Gool, L.: Diffir: Efficient diffusion model for image restoration. In: Proceedings of the IEEE/CVF International Conference on Computer Vision (ICCV) (2023)

\bibitem{Jianxiong_Xiao_Ehinger_Oliva_Torralba_2012}
Xiao, J., Ehinger, K.A., Oliva, A., Torralba, A.: Recognizing scene viewpoint using panoramic place representation. In: Proceedings of the IEEE Conference on Computer Vision and Pattern Recognition (CVPR) (2012)

\bibitem{yagi1999omnidirectional}
Yagi, Y.: Omnidirectional sensing and its applications. IEICE Transactions on Information and Systems (TOIS)  (1999)

\bibitem{yamazawa1993omnidirectional}
Yamazawa, K., Yagi, Y., Yachida, M.: Omnidirectional imaging with hyperboloidal projection. In: Proceedings of the IEEE/RSJ International Conference on Intelligent Robots and Systems (IROS) (1993)

\bibitem{yang2023rerender}
Yang, S., Zhou, Y., Liu, Z., Loy, C.C.: Rerender a video: Zero-shot text-guided video-to-video translation. In: Proceedings of the SIGGRAPH Asia 2023 Conference Papers (2023)

\bibitem{wang2023gpsr}
Yinhuai, W., Yujie, H., Jiwen, Y., Jian, Z.: Gan prior based null-space learning for consistent super-resolution. In: Proceedings of the AAAI Conference on Artificial Intelligence (AAAI) (2023)

\bibitem{Yoon_Chung_Wang_Yoon}
Yoon, Y., Chung, I., Wang, L., Yoon, K.J.: Spheresr: 360deg image super-resolution with arbitrary projection via continuous spherical image representation. In: Proceedings of the IEEE/CVF Conference on Computer Vision and Pattern Recognition (CVPR) (2022)

\bibitem{osrt_Yu_Wang_Cao_Li_Shan_Dong_2023}
Yu, F., Wang, X., Cao, M., Li, G., Shan, Y., Dong, C.: Osrt: Omnidirectional image super-resolution with distortion-aware transformer. In: Proceedings of the IEEE/CVF Conference on Computer Vision and Pattern Recognition (CVPR) (2023)

\bibitem{yu2023cross}
Yu, J., Zhang, X., Xu, Y., Zhang, J.: Cross: Diffusion model makes controllable, robust and secure image steganography. In: Proceedings of the Advances in Neural Information Processing Systems (NeurIPS) (2023)

\bibitem{yue2024resshift}
Yue, Z., Wang, J., Loy, C.C.: Resshift: Efficient diffusion model for image super-resolution by residual shifting. In: Advances in Neural Information Processing Systems (NeurIPS) (2023)

\bibitem{Zhang_2021_ICCV}
Zhang, K., Liang, J., Van~Gool, L., Timofte, R.: Designing a practical degradation model for deep blind image super-resolution. In: Proceedings of the IEEE/CVF International Conference on Computer Vision (ICCV) (2021)

\bibitem{zhang2018perceptual}
Zhang, R., Isola, P., Efros, A.A., Shechtman, E., Wang, O.: The unreasonable effectiveness of deep features as a perceptual metric. In: Proceedings of the IEEE Conference on Computer Vision and Pattern Recognition (CVPR) (2018)

\bibitem{zhang2024real}
Zhang, W., Li, X., Shi, G., Chen, X., Qiao, Y., Zhang, X., Wu, X.M., Dong, C.: Real-world image super-resolution as multi-task learning. In: Advances in Neural Information Processing Systems (NeurIPS) (2023)

\bibitem{zhang2022herosnet}
Zhang, X., Zhang, Y., Xiong, R., Sun, Q., Zhang, J.: Herosnet: Hyperspectral explicable reconstruction and optimal sampling deep network for snapshot compressive imaging. In: Proceedings of the IEEE/CVF Conference on Computer Vision and Pattern Recognition (CVPR) (2022)

\bibitem{WS_SSIM_Zhou_Yu_Ma_Shao_Jiang_2018}
Zhou, Y., Yu, M., Ma, H., Shao, H., Jiang, G.: Weighted-to-spherically-uniform ssim objective quality evaluation for panoramic video. In: Proceedings of the IEEE International Conference on Signal Processing (ICSP) (2018)

\end{thebibliography}

\appendix
\newpage
\addcontentsline{toc}{section}{Supp}






\section*{
\centering\Large
Supplementary Materials of ``OmniSSR: Zero-shot Omnidirectional Image 
Super-Resolution using Stable Diffusion Mode''
}


\section{Extra Experiments}
\label{sec:intro_ap}

\subsection{Ablation Studies}
\subsubsection{Ablation study of $\gamma$ on Gradient Decomposition (GD) correction}
According to the principle of GD correction, the super-resolution (SR) result in equirectangular projection (ERP) format $\mathbf{E}_{0|t}$ generated by StableSR~\cite{StableSR_Wang_Yue_Zhou_Chan_Loy_2023} can be further corrected to $\tilde{\mathbf{E}}_{0|t}=\mathbf{E}_{0|t}+\gamma\mathbf{A}^{\dagger}(\mathbf{E}_{init}-\mathbf{A}\mathbf{E}_{0|t})$, where $\gamma$ balances realness and fidelity. To improve the convergence of this gradient-based technique, we perform a grid search over different $\gamma$ values to obtain the best results, presented in Tab.~\ref{tab:ablation-gamma}. For an overall performance superiority, we choose $\gamma_l=0.5, \gamma_p=1, \gamma_e=1$.

\begin{figure}
    \centering
    \includegraphics[width=1.0\linewidth]{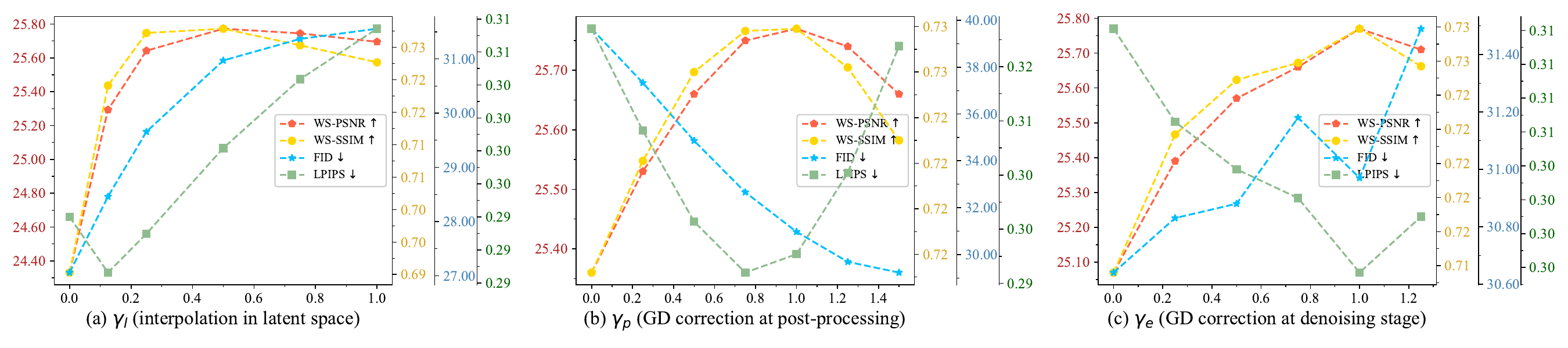}
    \caption{Visualization of different choices of $\gamma$. (a) $\gamma_p$ and $\gamma_e$ fixed, while adjusting $\gamma_l$; (b) $\gamma_e$ and $\gamma_l$ fixed, while adjusting $\gamma_p$; (c) $\gamma_p$ and $\gamma_l$ fixed, while adjusting $\gamma_e$.}
    \label{fig:vis-gamma}
\end{figure}

\begin{table}[h]
\centering
\scriptsize
\vspace{-3mm}
\caption{Ablation studies of hyper-parameter $\gamma$ in GD correction. $\gamma_p$ denotes $\gamma$ in post-processing stage, $\gamma_l$ denotes $\gamma$ in post-processing stage, $\gamma_e$ denotes $\gamma$ in post-processing stage. The best results are shown in \textbf{Bold}.}
\label{tab:ablation-gamma}
\begin{tabular}{c|c|c|cccc}
\toprule
$\gamma_p$ & $\gamma_l$ & $\gamma_e$ & WS-PSNR$\uparrow$ & WS-SSIM$\uparrow$ & FID$\downarrow$&LPIPS$\downarrow$ \\
\midrule
1&0&1 &24.33&0.6903&\textbf{27.05}&0.2925\\
\hline
1&0.25&1 &25.64&0.7272&29.66&\textbf{0.2912}\\
\hline
1&0.5&1 &\textbf{25.77}&\textbf{0.7279}&30.97&0.2977\\
\hline
1&0.75&1 &25.74&0.7253&31.37&0.3029\\
\hline
1&1&1 &25.69&0.7227&31.56&0.3067\\
\midrule

\midrule
0&0.5&1 &25.37&0.7172&39.64&0.3184\\
\hline
0.25&0.5&1 &25.53&0.7221&37.303&0.3090\\
\hline
0.5&0.5&1 &25.67&0.7260&34.86&0.3037\\
\hline
0.75&0.5&1 &25.75&0.7278&32.66&0.2960\\
\hline
1&0.5&1 &25.77&0.7279&30.97&0.2977\\
\hline
1.25&0.5&1 &25.74&0.7262&29.69&0.3052\\
\hline
1.5&0.5&1 &25.66&0.7230&29.22&0.3169\\
\midrule

\midrule
1&0.5&0 &25.07&0.7136&30.64&0.3121\\
\hline
1&0.5&0.25 &25.38&0.7217&30.83&0.3066\\
\hline
1&0.5&0.5 &25.56&0.7249&30.88&0.3037\\
\hline
1&0.5&0.75 &25.66&0.7259&31.18&0.3020\\
\hline
1&0.5&1 &25.77&0.7278&30.97&0.2977\\
\hline
1&0.5&1.25 &25.71&0.7257&31.49&0.3010\\

\bottomrule
\end{tabular}

\end{table}

\subsubsection{Ablation study of SR backbone}
We further conducted ablation studies on the selection of the SR backbone network to justify our choice of StableSR as the backbone and demonstrate the effectiveness of our proposed strategy at the same time. We selected the current state-of-the-art method in super-resolution work, SwinIR~\cite{liang2021swinir}, to compare its results with StableSR~\cite{StableSR_Wang_Yue_Zhou_Chan_Loy_2023}, which is shown in Tab.~\ref{tab:backbone}.

\begin{table}[h]
\scriptsize
\centering
\caption{Results of our proposed techniques on different backbones, StableSR, and SwinIR. Best results are shown in \textbf{Bold}.}
\label{tab:backbone}
\begin{tabular}{c|c|cccc}
\toprule
Backbone & Whether to use proposed techniques & WS-PSNR$\uparrow$ & WS-SSIM$\uparrow$ & FID$\downarrow$ & LPIPS$\downarrow$ \\
\hline
SwinIR~\cite{liang2021swinir}&$\times$&26.11&0.7821&27.11&0.2390\\
\hline
SwinIR~\cite{liang2021swinir}&$\checkmark$&27.89&0.8409&13.33&\textbf{0.1510}   \\
\hline
StableSR~\cite{StableSR_Wang_Yue_Zhou_Chan_Loy_2023}&$\checkmark$&\textbf{28.58}&\textbf{0.8540}&\textbf{13.01}&0.1575 \\
\bottomrule
\end{tabular}
\end{table}

Compared with SwinIR, StableSR significantly improves the fidelity and realness of reconstruction results. On the other hand, it also validates the effectiveness of our proposed Octadecaplex Tangent Information Interaction (OTII) and GD correction techniques on different backbones. Given its iterative updating and continuous correction nature, StableSR indeed has advantages over SwinIR's end-to-end reconstruction approach.

\subsection{Further Exploration of ERP$\leftrightarrow$TP Transformation}
\begin{figure*}[h]
	\scriptsize
	\centering
	\begin{tabular}{l}
		\hspace{-0.42cm}
		\hspace{-2mm}
		\begin{adjustbox}{valign=t}
			\begin{tabular}{cc}
				\includegraphics[width=0.5\linewidth]{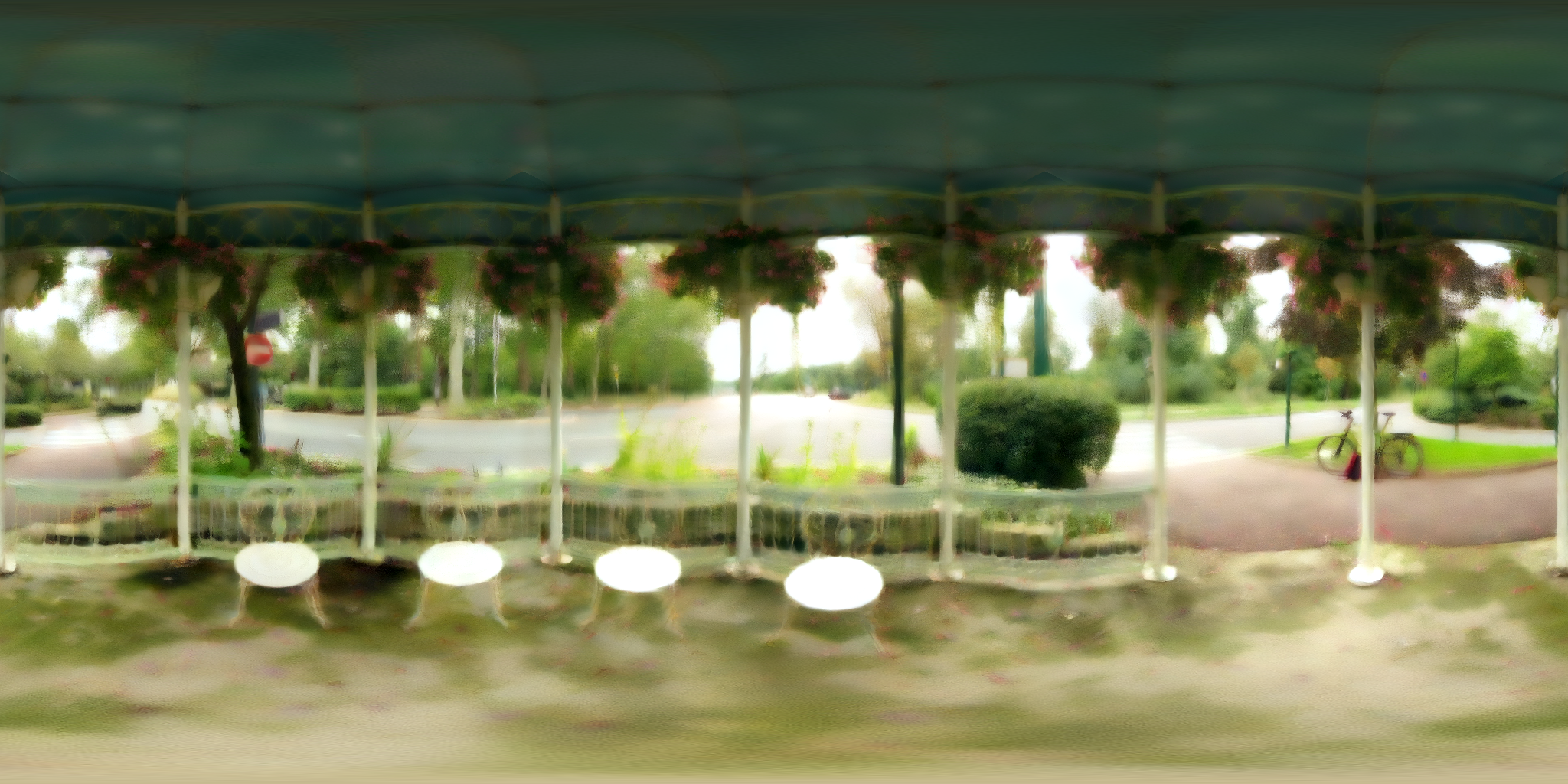} \hspace{-1mm} &
				\includegraphics[width=0.5\linewidth]{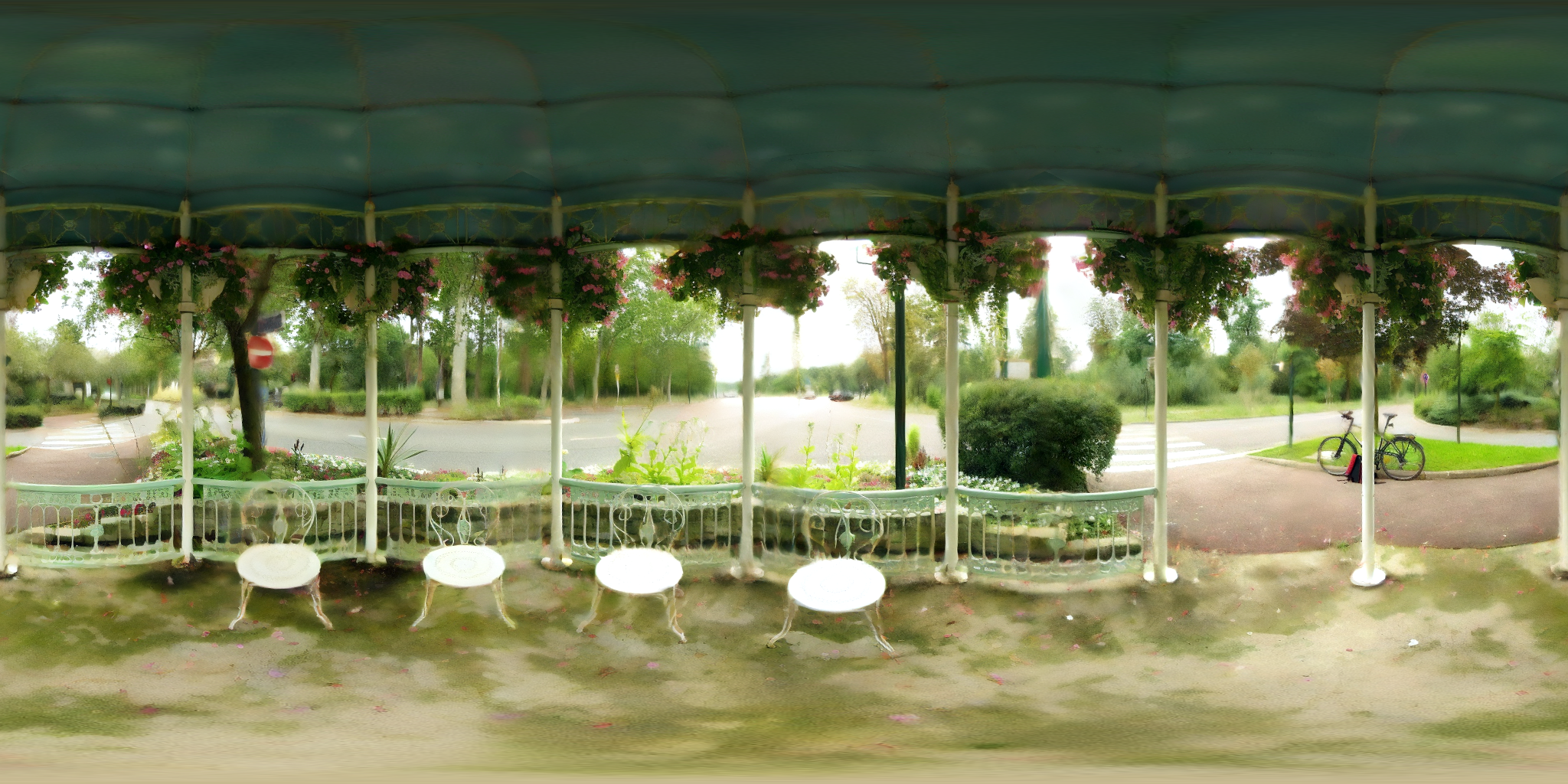} \hspace{-1mm} 
				\\
				On latent feature \hspace{-1mm} &
				On latent feature \hspace{-1mm} 
				\\ 
				(without pre-upsampling) \hspace{-1mm} &
				(with pre-upsampling) \hspace{-1mm} 
				\\
				\includegraphics[width=0.5\linewidth]{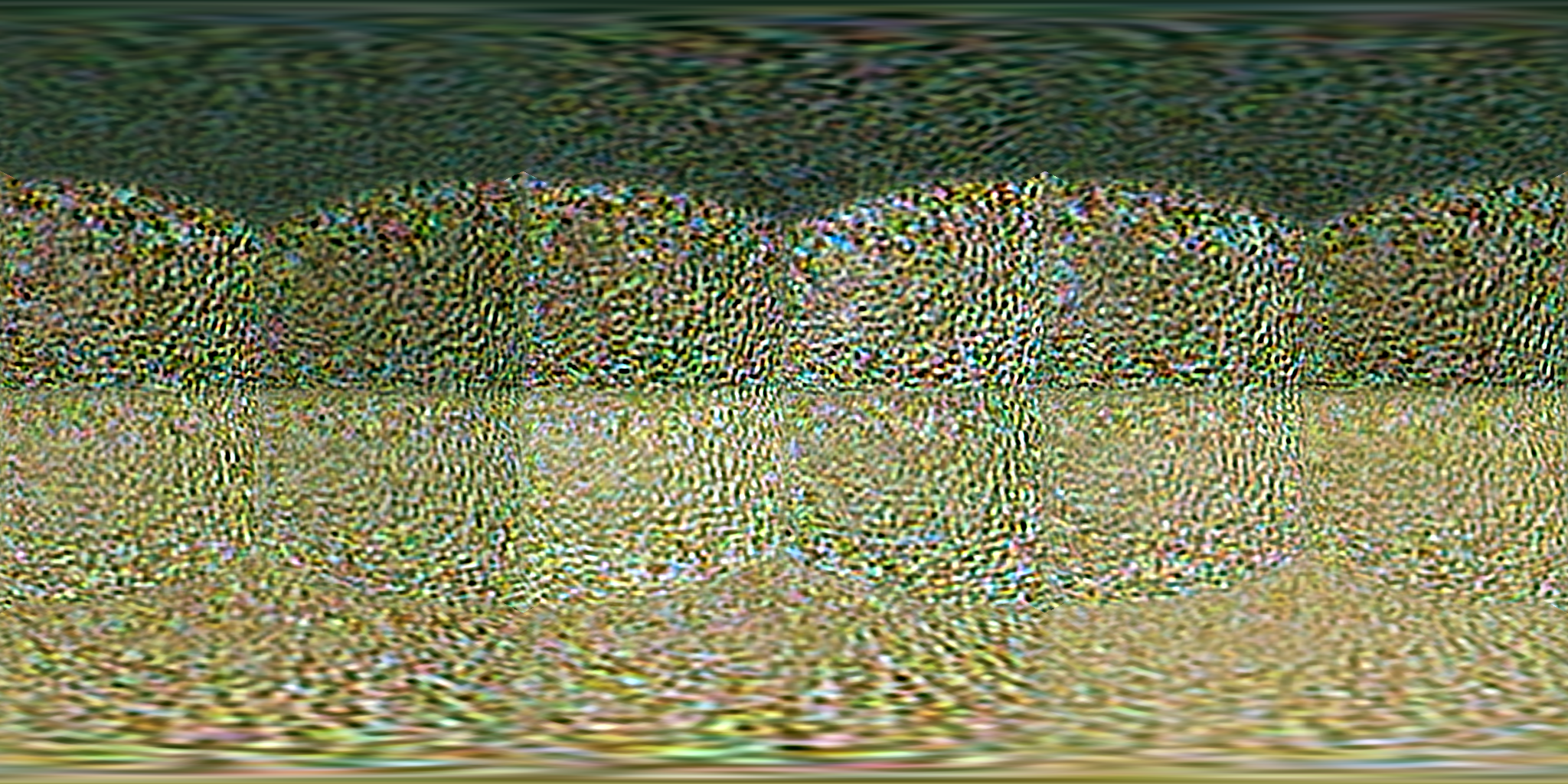} \hspace{-1mm} &
				\includegraphics[width=0.5\linewidth]{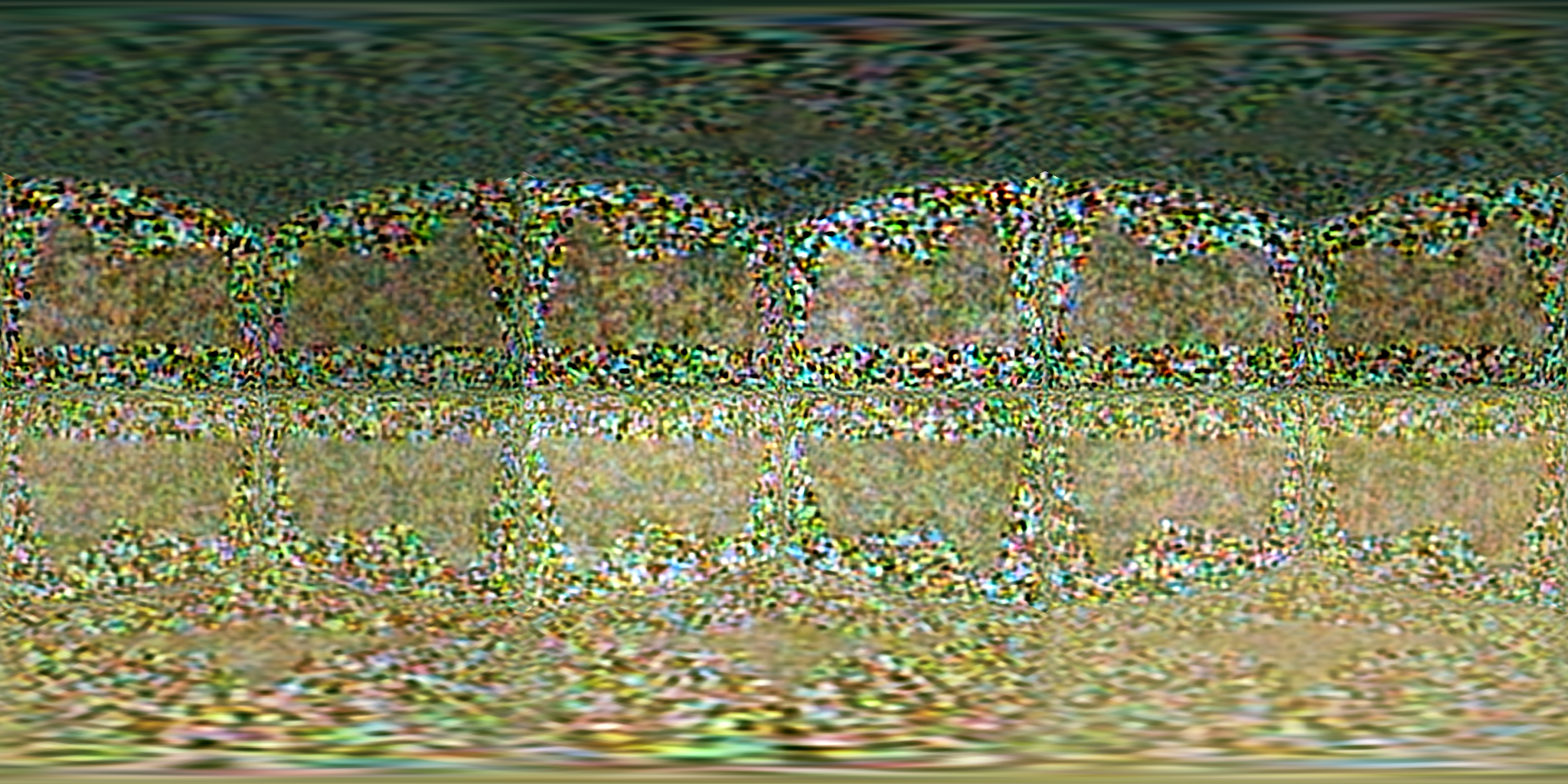} \hspace{-1mm}  
				\\ 
				On the latent noise \hspace{-1mm} &
				On the latent noise \hspace{-1mm} 
				\\
				(without pre-upsampling) \hspace{-1mm} &
				(with pre-upsampling) \hspace{-1mm} 
			\end{tabular}
		\end{adjustbox}
        \end{tabular}
	\caption{Visualized comparison of projection transformations on latent image feature and latent noise. Zoom in for details.}
	\label{fig:proj}
\end{figure*}

A simple question arises: can we perform ERP$\leftrightarrow$TP\footnote{TP denotes tangent projection.} transformation in the latent space, thus avoiding the need to transform intermediate results between image and latent space repeatedly? To answer this question, we made two attempts without Stable Diffusion (SD) encoder and decoder during each denoising step. GD correction is also not used in this section.

1) \textbf{Projection transformations on latent feature $z_0$:} In this experiment, we focus on the impact of projection transformation on image features in the latent space, so here we do not involve the denoising process. Therefore, we first transformed the ground truth ERP image $\mathbf{E}_0$ to $m$ TP images $\{\mathbf{x}_0^{(i)}\}_{i=1,...,m}$ through ERP$\rightarrow$TP. Then, we sequentially obtain the latent TP image features in the latent space: 
\begin{equation}
    \begin{aligned}
        &\mathbf{z}_0^{(i)} = \mathcal{E}(\mathbf{x}_0^{(i)}), i=1,...,m.
    \end{aligned}
\end{equation} 
Next, we perform TP$\rightarrow$ERP$\rightarrow$TP on $\mathbf{z}_0^{(i)}$ to obtain $\hat{\mathbf{z}}_0^{(i)}$ and decode them to TP image as follows: 
\begin{equation}
    \begin{aligned}
        &\hat{\mathbf{x}}_0^{(i)} = \mathcal{D}(\hat{\mathbf{z}}_0^{(i)}), i=1,...,m.
    \end{aligned}
\end{equation} 
Finally, the decoded TP image $\hat{\mathbf{x}}_0^{(i)}$ are transformed by TP$\rightarrow$ERP to get $\hat{\mathbf{E}}_0$. 

2) \textbf{Projection transformations on latent noise $\boldsymbol{\epsilon}_t^{(i)}$:} In this experiment, we focus on the impact of projection transformation on the noise $\boldsymbol{\epsilon}_t^{(i)}$. We transform the low-resolution ERP image to TP images and feed the latter into StableSR pipeline. At each sampling step, we directly perform TP$\rightarrow$ERP$\rightarrow$TP transformation on the predicted noise $\{\boldsymbol{\epsilon}_t^{(i)}\}_{i=1,...,m}$ to get $\{\hat{\boldsymbol{\epsilon}}_t^{(i)}\}_{i=1,...,m}$, and using $\hat{\boldsymbol{\epsilon}}_t^{(i)}$ for following denoising.


In the two experiments above, we also present the effects of using and not using pre-upsampling in the TP$\rightarrow$ERP$\rightarrow$TP transformation process, respectively. We illustrate the visual results of $\hat{\mathbf{E}}_0$, using the 0000.png in image ODI-SR testset as an example in Fig.~\ref{fig:proj}. When \textbf{performing projection transformations on latent feature $z_0$}, the decoded images exhibit severe blurring. Although using pre-upsampling in the TP$\rightarrow$ERP$\rightarrow$TP process can alleviate the blurriness to some extent and present clearer image content in certain areas, the overall image quality remains poor. In the experiment involving \textbf{projection transformations on latent noise $\boldsymbol{\epsilon}_t^{(i)}$}, it can be observed that regardless of whether pre-upsampling strategy is used or not, the super-resolved images suffer from significant damage. This may be attributed to the SD encoder's spatial downsampling at $\times$8 scale, compressing image pixels within an 8$\times$8 patch into a single latent pixel. Projection transformations, on the other hand, operate at the image pixel level with fine granularity. Applying such fine-grained operations directly to latent pixels can greatly disrupt the original image structure. Therefore, projection transformations related to ODIs should be performed in image space rather than in the latent space mapped by the SD Variational Auto Encoder (VAE).

\subsection{Exploration of SD Encoder and Decoder}
\begin{figure}
    \centering
    \includegraphics[width=1\linewidth]{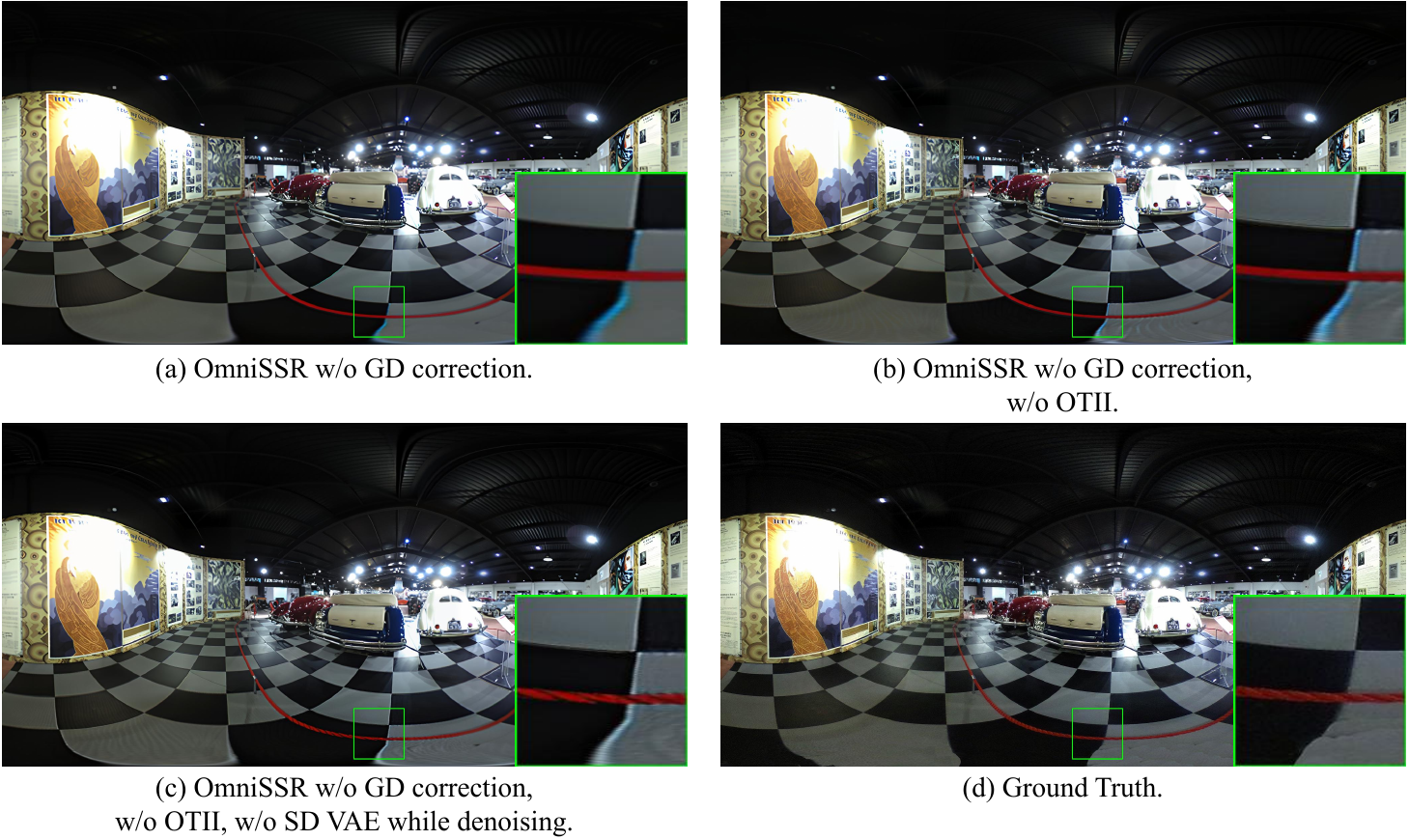}
    \caption{Phenomenon and causes of color shift: By progressively removing different components of OmniSSR (a)(b)(c), we ultimately discovered that the color shift in the super-resolution results disappears again after removing the SD VAE used in the denoising step. This indicates the potential risk of color shift associated with frequent usage of SD VAE during denoising.}
    \label{fig:corlor-shift}
\end{figure}

During the ablation study, we observed that OmniSSR, when GD correction is removed while OTII is retained, demonstrates improved fidelity (e.g., WS-PSNR, WS-SSIM) and deteriorated realness (e.g., FID, LPIPS) compared to the original StableSR model. Upon examining the outputs of the ablation model under this configuration, significant color shift issues were identified, as depicted in Fig.~\ref{fig:corlor-shift}(a). 

We initially suspected that this color shift stemmed from \textbf{the utilization of the SD VAE} before and after OTII in each denoising step. To validate this hypothesis, we conducted a visual comparison experiment using image 0006.png from the ODI-SR testset as an example. It can be observed that even when GD correction and OTII are successively removed, as illustrated in Fig.~\ref{fig:corlor-shift}(a)(b), the color shift persists. It is only when we eliminate the repeated usage of SD VAE in each denoising step that the color at the boundary of black and white tiles returns to normal, as shown in Fig.~\ref{fig:corlor-shift}(c). Ground truth reference can be seen in Fig.~\ref{fig:corlor-shift}(d). This phenomenon of color shift indicates the potential problem caused by frequently using SD VAE.

\subsection{The Global Continuity of ODIs}
\begin{figure}
    \centering
    \includegraphics[width=1\linewidth]{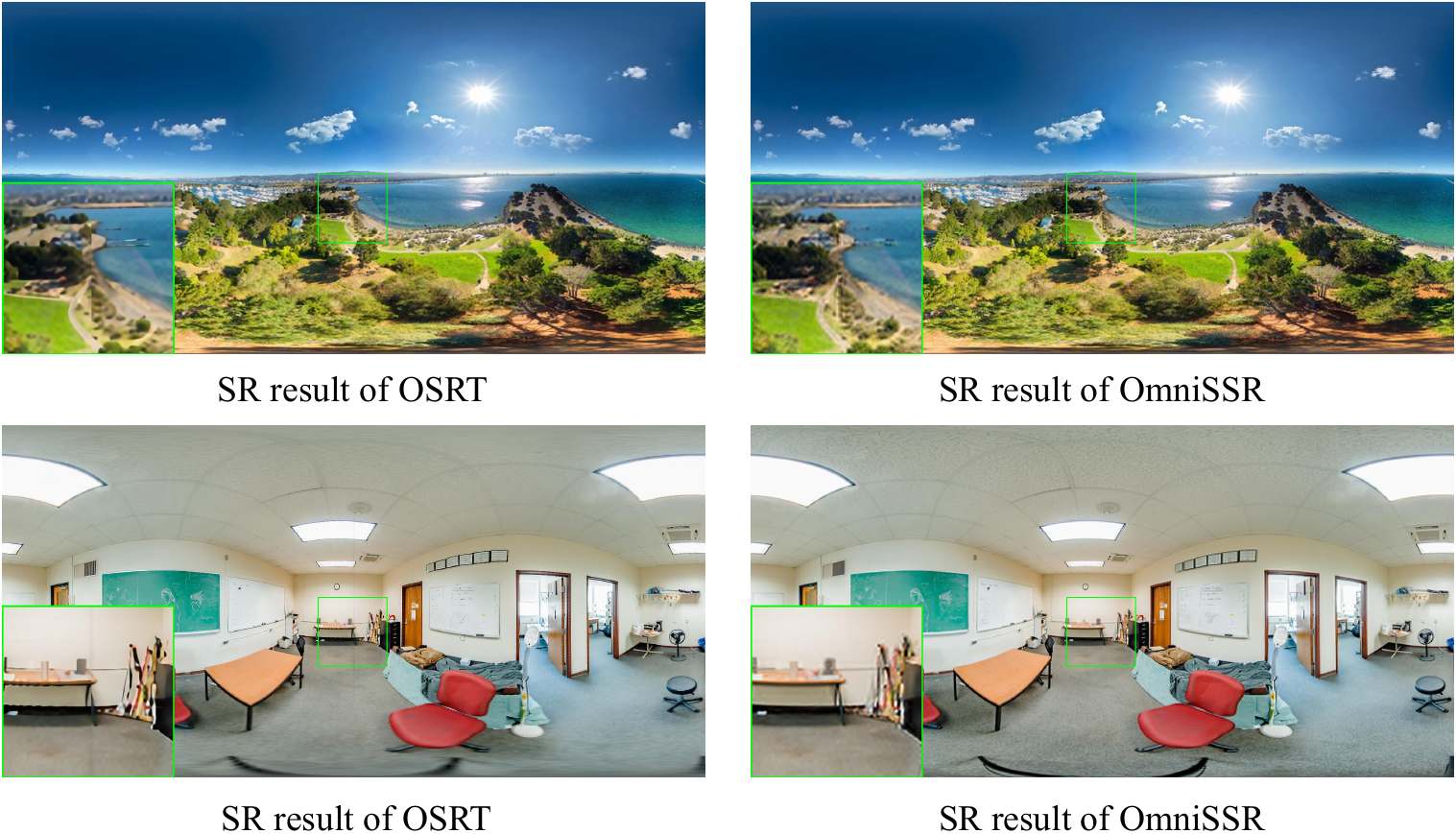}
    \caption{Continuity of left and right part of SR results on OSRT and our proposed OmniSSR. It is shown that OSRT suffers from serious artifacts and bad continuity. All ERP images have been rotated by 180 degrees to stitch the left and right sides. (Upper image: 0039 of ODI-SR test set, lower image: 0015 of SUN test set.)}
    \label{fig:continuity}
\end{figure}

The existing ODISR methods directly perform SR on ERP images, resulting in the discontinuity between the left and right sides~\cite{Cao_2023_CVPR}. Our proposed OTII treats TP images as the direct input for the network. Besides facilitating the transfer use of existing planar image-specific diffusion models, it also effectively considers the omnidirectional characteristics of ODIs. We selected some visualization results of OSRT~\cite{osrt_Yu_Wang_Cao_Li_Shan_Dong_2023} and OmniSSR, focusing on the continuity near the left and right sides of the ERP. As shown in Fig.~\ref{fig:continuity}, OSRT exhibits poor continuity between the left and right sides of the ERP, while OmniSSR naturally inherits the advantage of TP images in seamlessly spanning different areas of the ERP.


\subsection{Time Consumption}
The inference runtime of different methods are compared as follows. Considering fair comparison, we use the default settings referred to in corresponding papers. The diffusion sampling steps for OmniSSR are 200, DDRM~\cite{kawar2022denoising} 100, and PSLD~\cite{Rout_Raoof_Daras_Caramanis_Dimakis_Shakkottai_2023} 1000.\footnote{We have tried to use the same sampling accelerate strategy in DDRM, but get bad restored results.} All experiments are conducted on a single NVIDIA 3090Ti GPU.
\begin{table}[h]
\centering
\scriptsize
\caption{Time consumption of OmniSSR and other SR methods.}
\label{tab:time}
\begin{tabular}{c|c}
\toprule
Method & Runtime per ERP image (s)$\downarrow$\\
\hline
SwinIR~\cite{liang2021swinir}&0.87\\ 
\hline
OSRT~\cite{osrt_Yu_Wang_Cao_Li_Shan_Dong_2023}&1.44\\
\hline
DDRM&711.95\\
\hline
PSLD&6720.87\\
\hline
OmniSSR (Ours)&726.19\\

\bottomrule
\end{tabular}

\end{table}
\section{Theoretical Discussion}

In this section, we provide a simple theoretical discussion of our proposed GD correction technique, explaining why a single step of GD would also work and obtain better results.

Take the update step in GD correction as an example, let us first re-examine this step:
\begin{equation}
    \tilde{\mathbf{E}}_{0|t}=\mathbf{E}_{0|t}+\gamma_e \mathbf{A}^{\dagger}(\mathbf{E}_{init}-\mathbf{A}\mathbf{E}_{0|t}),
\end{equation}
where $\gamma_e \mathbf{A}^{\dagger}(\mathbf{E}_{init}-\mathbf{A}\mathbf{E}_{0|t})$ is the gradient of fidelity term $||\mathbf{E}_{init}-\mathbf{A}\mathbf{E}_{0|t}||_F$, and $\gamma_e=2\times \alpha \text{ (learning rate)}$.

An obvious and direct question is: why did we perform only a single update step rather than multiple steps? Through the following analysis, we will demonstrate that, in this context, multi-step gradient descent and single-step are essentially equivalent, with the number of steps being governed by the coefficient $\gamma_e$.

\textbf{Analysis} Suppose we take multiple steps in GD correction and are taking step $k$ to $k-1$. As $\tilde{\mathbf{E}}_{0|t}^{(k)}$ can be represented via $\tilde{\mathbf{E}}_{0|t}^{(k-1)}$ in linear form, we can use $\tilde{\mathbf{E}}_{0|t}^{(0)}$ to express $\tilde{\mathbf{E}}_{0|t}^{(k)}$, and $\tilde{\mathbf{E}}_{0|t}^{(0)}$ only has linear coefficients composed of $\gamma_e$, $\mathbf{A}$ and $\mathbf{A}^{\dagger}$. 
Thus for fixed $\gamma_e$, there is no difference between one step and multiple steps of GD correction.
For adaptive $\gamma_e$, it is also obvious that $\tilde{\mathbf{E}}_{0|t}^{(k)}$ can be represented via $\tilde{\mathbf{E}}_{0|t}^{(0)}$ with linear transforms and different $\gamma_e$. Thus for a better trade-off between performance and inference time, we turn to use \textbf{one} step of GD correction.


%
%

\end{document}